\documentclass[twoside,11pt]{article}

\usepackage{subcaption}
\usepackage[preprint]{jmlr2e}

\usepackage{amsmath}
\usepackage{mathtools}
\usepackage{commath}
\usepackage{enumitem}
\usepackage{bbm}

\newcommand{\1}{\mathbf{1}}
\newcommand{\tp}{\top}

\newcommand{\SDP}{\text{SDP}}
\newcommand{\CSDP}{\text{CSDP}}
\newcommand{\la}{\langle}
\newcommand{\ra}{\rangle}

\newcommand{\me}{\mathrm{e}}

\DeclareMathOperator{\prob}{P}
\DeclareMathOperator{\E}{E}
\DeclareMathOperator{\V}{Var}
\DeclareMathOperator{\G}{\mathcal{G}}
\DeclareMathOperator{\Tr}{Tr}

\DeclareMathOperator*{\argmax}{arg\,max}

\jmlrheading{[volume]}{2022}{[pages]}{8/22}{[date published]}{[paper id]}{Junda Sheng and Thomas Strohmer}

\ShortHeadings{Semi-Supervised Clustering of Sparse Graphs}{Sheng and Strohmer}
\firstpageno{1}

\begin{document}
\title{Semi-Supervised Clustering of Sparse Graphs: \\ Crossing the Information-Theoretic Threshold}

\author{\name Junda Sheng \email sheng@math.ucdavis.edu \\
       \addr Department of Mathematics\\
       University of California\\
       Davis, CA 95616-5270, USA
       \AND
       \name Thomas Strohmer \email strohmer@math.ucdavis.edu \\
       \addr Department of Mathematics and Center of Data Science and Artificial Intelligence Research\\
       University of California\\
       Davis, CA 95616-5270, USA}

\editor{}

\maketitle

\begin{abstract}%
The stochastic block model is a canonical random graph model for clustering and community detection on network-structured data. Decades of extensive study on the problem have established many profound results, among which the phase transition at the Kesten-Stigum threshold is particularly interesting both from a mathematical and an applied standpoint. It states that no estimator based on the network topology can perform substantially better than chance on sparse graphs if the model parameter is below a certain threshold.
Nevertheless, if we slightly extend the horizon to the ubiquitous semi-supervised setting, such a fundamental limitation will disappear completely. We prove that with an arbitrary fraction of the labels revealed, the detection problem is feasible throughout the parameter domain. Moreover, we introduce two efficient algorithms, one combinatorial and one based on optimization, to integrate label information with graph structures. Our work brings a new perspective to the stochastic model of networks and semidefinite program research.
\end{abstract}


\section{Introduction}
Clustering has long been an essential subject of many research fields, such as machine learning, pattern recognition, data science, and artificial intelligence. In this section, we include some background information on its general setting and the semi-supervised approach.

\subsection{Clustering on Graphs}
The basic task of \textit{clustering} or \textit{community detection} in its general form is, given a (possibly weighted) graph, to partition its vertices into several densely connected groups with relatively weak external connectivity. This property is sometimes also called assortativity. Clustering and community detection are central problems in machine learning and data science with various applications in scientific research and industrial development. A considerable amount of data sets can be represented in the form of a network that consists of interacting nodes, and one of the first features of interest in such a situation is to understand which nodes are “similar”, as an end or as a preliminary step towards other learning tasks. Clustering is used to find genetically similar sub-populations \citep{10.3389/fgene.2014.00204}, to segment images \citep{868688}, to study sociological behavior \citep{doi:10.1073/pnas.012582999}, to improve recommendation systems \citep{1167344}, to help with natural language processing \citep{doi:10.1073/pnas.1221839110}, etc. Since the 1970s, in different communities like social science, statistical physics, and machine learning, a large diversity of algorithms have been developed such as:
\begin{itemize}
    \item Hierarchical clustering algorithms \citep{hierarchy} build a hierarchy of progressive communities, by either recursive aggregation or division.
    \item Model-based statistical methods, including the celebrated EM clustering algorithm proposed in \citep{https://doi.org/10.1111/j.2517-6161.1977.tb01600.x}, fit the data with cluster-exhibiting statistical models.
    \item Optimization approaches identify the best cluster structures regarding carefully designed cost functions, for instance, minimizing the cut \citep{HARTUV2000175} and maximizing the Girvan-Newman modularity \citep{Newman2004FindingAE}.
\end{itemize}

Multiple lines of research intersect at a simple random graph model, which appears under many different names. In the machine learning and statistics literature around social networks, it is called stochastic block model (SBM) \citep{Holland1983StochasticBF}, while it is known as the planted partition model \citep{715914} in theoretical computer science and referred to as inhomogeneous random graph model \citep{10.5555/1276871.1276872} in the mathematics literature. Moreover, it can also be interpreted as a spin-glass model \citep{Decelle2011AsymptoticAO}, a sparse-graph code \citep{Abbe2015CommunityDI}, a low-rank random matrix model \citep{959929}, and more. 

The essence of SBM can be summarized as follows: Conditioned on the vertex labels, edges are generated independently and the probability only depends on which clusters the pairs of vertices belong to. We consider its simplest form,  namely the symmetric SBM consisting of two blocks, also known as the planted bisection model.
    
\begin{definition}[Planted bisection model]\label{PBM}
For $n\in \mathbb{N}$ and $p,q \in (0,1)$, let $\G(n,p,q)$ denote the distribution over graphs with $n$ vertices defined as follows. The vertex set is partitioned uniformly at random into two subsets $S_1, S_2$ with $|S_i|=n/2$. Let $E$ denote the edge set. Conditional on this partition, edges are included independently with probability
\begin{equation}
    \prob\left((i,j)\in E | S_1, S_2\right)=
    \begin{cases}
    p & \text{~if~} \{i,j\} \subseteq S_1 \text{~or~} \{i,j\} \subseteq S_2, \\
    q & \text{~if~} i \in S_1, j \in S_2 \text{~or~} i \in S_2, j \in S_1.
    \end{cases}
\end{equation}
\end{definition}

Note that if $p=q$, the planted bisection model is reduced to the so-called Erdős–Rényi random graph where all edges are generated independently with the same probability. Hence there exists no cluster structure. But if $p \gg q$, a typical graph will have two well-defined clusters. The scale of $p$ and $q$ also plays a significant role in the resulting graph, which will be discussed in detail later. They govern the amount of signal and noise in the graph's generating process. As the key parameters that researchers work with, they depict various regimes and thresholds.

The SBMs generate labels for vertices before the graph. The ground truth allows us to formally discuss the presence of community structures and measure the performance of algorithms in a meaningful way. It also supplies a natural basis to rigorously define the semi-supervised clustering problem. But as a parametrized statistical model, one can only hope that it serves as a good fit for the real data. Although not necessarily a realistic model, SBM provides us with an insightful abstraction and captures some of the key phenomena \citep{MosselNS15,  Chen2016StatisticalComputationalTI, Banks2016InformationtheoreticTF, Abbe2016ExactRI,ASgen}.

Given a single realization of a graph $G$, our goal is to recover the labels $x$, up to a certain level of accuracy. Formally, the ground truth of the underlying community structure is encoded using the vector $x \in \{+1,-1\}^n$, with $x_i = +1$ if $i\in S_1$, and $x_i = -1$ if $i\in S_2$. An estimator is a map $\hat{x}: G_n \to \{+1,-1\}^n$ where $G_n$ is the space of graphs over $n$ vertices. We define the \textit{Overlap} between an estimator and the ground truth as
\begin{equation}
     \text{Overlap}(x,\hat{x}(G))=\frac{1}{n}| \langle x,\hat{x}(G) \rangle |.
\end{equation}

{\em Overlap} induces a measure on the same probability space as the model, which represents how well an (unsupervised) estimator performs on the recovery task. To intuitively interpret the result, we put requirements on its asymptotic behavior, which takes place with high probability as $n\to\infty$.
\begin{definition}
Let $G \sim \G(n,p,q)$. The following recovery requirements are solved if there exists an algorithm that takes $G$ as an input and outputs $\hat{x}=\hat{x}(G)$ such that
\begin{itemize}
\item Exact recovery: $\prob\{\text{Overlap}(x,\hat{x}(G))=1\}=1-o(1)$ 
\item Weak recovery: $\prob\{\text{Overlap}(x,\hat{x}(G)) \geq \Omega(1) \}=1-o(1)$
\end{itemize}
\end{definition}
In other words, exact recovery requires the entire partition to be correctly identified. Weak recovery only asks for substantially better performance than chance. In some literature, exact recovery is simply called recovery. Weak recovery is also called detection since as long as one can weakly recover the ground truth, there must exist a community structure. 

Note that if $G$ is an Erdős–Rényi random graph ($p = q$) then the overlap will be $o_p(1)$ for all estimators. This can be seen by noticing that $x$ and $G$ are independent in this setting and then applying Markov's inequality. This has led to two additional natural questions about SBMs. On the one hand, we are interested in the distinguishability (or testing): is there a hypothesis test to distinguish a random graph generated by the Erdős–Rényi model (ERM) from a random graph generated by the SBM, which succeeds with high probability? On the other hand, we can ask about the model learnability (or parameter estimation): assuming that $G$ is drawn from an SBM ensemble, is it possible to obtain a consistent estimator for the parameters ($p,q$)? Although each of these questions is of independent interest, for symmetric SBMs with two symmetric communities (planted bisection model) the following holds \citep{Abbe2017CommunityDA}:
\begin{equation}
    \text{learnability}\iff\text{weak recovery}\iff\text{distinguishability}.
\end{equation}
Such equivalence benefits our understanding of the model in turn. For example, direct analysis of weak recovery leads to the converse of phase transition theory \citep{MosselNS15}. The achievability of the phase transition threshold \citep{Massouli2014CommunityDT} is proved by counting non-backtracking walks on the graph which gives consistent estimators of parameters. In the recent work \citep{montanari2015semidefinite}, hypothesis testing formulation is studied.

SBMs demonstrate the `fundamental limits' of clustering and community detection as some necessary and sufficient conditions for the feasibility of recovery, information-theoretically or computationally. Moreover, they are usually expressed in the form of \textit{phase transition}. Sharp transitions exist in the parameter regimes between phases where the task is resolvable or not. For example, when the average degree grows as $\log n$, if the structure is sufficiently obvious then the underlying communities can be exactly recovered \citep{doi:10.1073/pnas.0907096106}, and the threshold at which this becomes possible has also been determined \citep{Abbe2016ExactRI}. Above this threshold, efficient algorithms exist \citep{Agarwal2015MultisectionIT, Abbe2015CommunityDI, Perry2017ASP, Deng2021StrongCG} that recover the communities exactly, labeling every vertex correctly with high probability; below this threshold, exact recovery is information-theoretically impossible. \citep{10.5555/3157096.3157205, JMLR:v18:16-245} studied a slightly weaker requirement that allows a vanishing proportion of misclassified nodes and proposed algorithms that achieve the optimal rates for various models. In contrast, we study a rather different regime where the reasonable question to ask is whether one can guarantee a misclassification proportion strictly less than $\frac{1}{2}$.

We acknowledge that SBMs in general are theoretical constructs that can not capture all the complicated situations emerging with real-life networks. However, it succeeds in modeling the simplest situation where a community structure can be quantitatively described. In practice, instead of applying the model directly, it is advisable to initially conduct some heuristic and hierarchical analyses on the graph and see if the problem can be effectively reduced to those simple cases. For the graphs violating the model assumptions, e.g.,\ a sparse graph dominated by cliques, alternative and tailored modeling may be necessary.
 
\subsection{Sparse Regime and Kesten-Stigum Threshold} \label{topo}
In the sparse case where the average degree of the graph is $O(1)$, it is more difficult to find the clusters and the best we can hope for is to label the vertices with nonzero correlation or mutual information with the ground truth, i.e.,\ weak recovery. Intuitively, we only have access to a constant amount of connections about each vertex. The intrinsic difficulty can be understood from the topological properties of the graphs in this regime. The following basic results are derived from \citep{Erdos1984OnTE}:
\begin{itemize}
    \item For $a, b > 0$, the planted bisection model $\G(n,\frac{a \log n}{n},\frac{b \log n}{n})$ is connected with high probability if and only if $\frac{a+b}{2}>1$.
    \item $\G(n,\frac{a}{n},\frac{b}{n})$ has a giant component (i.e.,\ a component of size linear in $n$) with high probability if and only if $d\coloneqq\frac{a+b}{2}>1$.
\end{itemize}
The graph will only have vanishing components if the average degree is too small. Therefore, it is not possible to even weakly recover the labels. But we will see in the next section that semi-supervised approaches amazingly piece the components together with consistent labeling.  

Although it is mathematically challenging to work in the sparse regime, real-world data are likely to have bounded average degrees. \citep{Leskovec} and \citep{Strogatz:2001wc} studied a large collection of benchmark data sets, including power transmission networks, website link networks, and complex biological systems, which had millions of nodes with an average degree of no more than 20. For instance, the LinkedIn network they studied had approximately seven million nodes, but only 30 million edges.

The phase transition for weak recovery or detection in the sparse regime was first conjectured in the paper by Decelle, Krzakala, Moore, Zdeborová \citep{Decelle2011AsymptoticAO}, which sparked the modern study of clustering and SBMs. Their work is based on deep but non-rigorous insights from statistical physics, derived with the cavity method (a.k.a.\ belief
propagation). Since then, extensive excellent research has been conducted to understand this fundamental limit, e.g.,\ \citep{MosselNS15, Massouli2014CommunityDT, Mossel2018APO, Abbe2020GraphPA}. A key result is the following theorem.

\begin{theorem}\label{KS}[Kesten-Stigum threshold]
Let $\G(n, a/n, b/n)$ be a symmetric SBM with two balanced clusters and $a, b = O(1)$. The weak recovery problem is solvable and efficiently so, if and only if $(a-b)^2 > 2(a + b)$.
\end{theorem}

In particular, if we denote the probability measures induced by the ERM $\G(n, \frac{a+b}{2n}, \frac{a+b}{2n})$ and the SBM $\G(n, \frac{a}{n}, \frac{b}{n})$ by $P_n^{(0)}$ and $P_n^{(1)}$ correspondingly, they are mutually contiguous, that is for any sequence of events $\{E_n\}$'s, $P_n^{(0)}(E_n) \to 0$ if, and only if, $P_n^{(1)}(E_n) \to 0$.

Conventionally, \emph{signal-to-noise ratio (SNR)} is defined as the following,
\begin{equation}
\text{SNR} \coloneqq (a-b)^2/[2(a+b)].
\end{equation}
It is worth noting that we only quoted the KS threshold for the two communities case ($k = 2$). For sufficiently large $k$, namely $k \geq 5$, there is a `hard but detectable' area where the weak recovery is information-theoretically possible, but computationally hard \citep{ASgen, Banks2016InformationtheoreticTF}. This gap between the KS threshold and the information-theoretic (IT) threshold only shows up in the constant degree regime, making it a fertile ground for studying the fundamental tradeoffs in community detection. We focus on the cardinal case, symmetric
SBM with two balanced clusters, where two thresholds coincide and a semi-supervised approach crosses them together.

\begin{figure}[ht]
\begin{subfigure}{0.3\textwidth}
\centering\includegraphics[width=\textwidth]{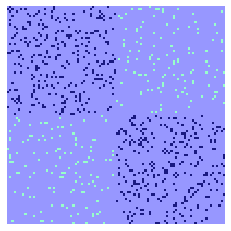}
\end{subfigure}
\hspace*{\fill}
\begin{subfigure}{0.3\textwidth}
\centering\includegraphics[width=\textwidth]{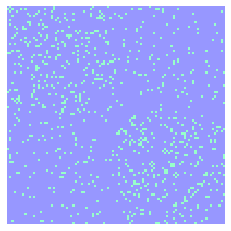}
\end{subfigure}
\hspace*{\fill}
\begin{subfigure}{0.3\textwidth}
\centering\includegraphics[width=\textwidth]{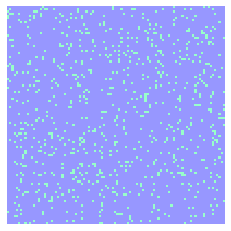}
\end{subfigure}
\hspace*{\fill} 
\caption{The left image represents the adjacency matrix of one realization of $\G (100, 0.12, 0.05)$, where the detection is theoretically possible. Yet the data is given non-colored (middle) and also non-ordered (right).}
\end{figure}

The terminology `KS threshold' can be traced back to the work of Kesten and Stigum concerning the reconstruction of infinite rooted trees in 1966 \citep{KS}. The problem consists of broadcasting the root label of a tree with a fixed degree $c$ down to its leaves and trying to recover it from the leaves at a large depth. We start with drawing the root label uniformly in $\{0,1\}$. Then, in a top-down manner, we independently label every child the same as its parent with probability $1-\epsilon$ and the opposite as its parent otherwise. Let $x^{(t)}$ denote the labels at depth $t$ in this tree with $t = 0$ being the
root. We say the reconstruction is solvable if $\lim_{t\to \infty} \E|\E(x^{(0)}|x^{(t)}) - 1/2| > 0$ or, equivalently, $\lim_{t\to\infty} I(x^{(0)}; x^{(t)}) > 0$, where $I$ is the mutual information. Although it was shown in the original paper, reconstruction is solvable when $c(1-2\epsilon)^2 > 1$, the non-reconstruction was proved 30 years later, namely, it is not solvable if $c(1-2\epsilon)^2 \leq 1$ \citep{Bleher1995OnTP, Evans}. Based on that finding, Mossel, Neeman, and Sly proved the converse part of Theorem \ref{KS} by coupling the local neighborhood of an SBM vertex with a Gaton-Watson tree with
a Markov Process \citep{MosselNS15}. Inspired by this elegant approach, we propose our `census method' to solve the semi-supervised clustering problem, and we will see in Section \ref{Census method} how it works by amplifying revealed information with tree-like neighborhoods.

\subsection{Basic Algorithms}
Information-theoretic bounds can provide the impossibility side of phase transitions, but we still need specific efficient algorithms for the achievability side. One straightforward approach is the spectral method. Under Definition \ref{PBM}, let $A$ be the adjacency matrix of the graph $G \sim \G(n,a/n,b/n)$, $a>b$. Up to reordering indices, its expectation is a block matrix except for the diagonal,
\begin{equation}
\E A \approx \frac{1}{n} \begin{pmatrix}
a & b\\
b & a
\end{pmatrix}\otimes I_{n/2 \times n/2},
\end{equation}
which has three eigenvalues, $(a+b)/n > (a-b)/n>0$. $0$ has multiplicity $n-2$ and the eigenvector associated with the second largest eigenvalue is $\left(\begin{smallmatrix}\1_{n/2}\\-\1_{n/2}\end{smallmatrix}\right)$ which is consistent with the ground truth of the labels. However, we do not observe the expected adjacency matrix. Instead, we only have access to one realization of the model. In modern terms, community detection is a `one-shot learning' task. But one can still hope that $A - \E A$ is small and the second eigenvector of $A$ gives a reasonable estimator. For example, denoting the ordered eigenvalues of $\E A$ and $A$ as $\{\lambda_i\}$'s and $\{\hat{\lambda}_i\}$'s respectively, the Courant-Fischer-Weyl min-max principle implies 
\begin{equation}
|\hat{\lambda}_i - \lambda_i | \leq \|A - \E A\|_{\text{op}}, \qquad i = 1,\dots,n. 
\end{equation}
Recall that the operator norm of a symmetric matrix $M$ is $\|M\|_{\text{op}} = \max(\xi_1(M), -\xi_n(M))$ where $\xi_i(M)$ denotes the $i$-th largest eigenvalue of $M$. If one can bound $\|A - \E A\|_{\text{op}}$ by half of the least gap between the three eigenvalues mentioned above, the order will be preserved. Then the Davis-Kahan theorem guarantees the eigenvectors are correlated. Namely, if $\theta$ denotes the angle between the second eigenvectors (spectral estimator and ground truth), we have
\begin{equation}
\sin \theta \leq \|A - \E A\|_{\text{op}}/ \min\{\|\lambda_i - \lambda_2\|/2 : i\neq 2\}.
\end{equation}
Thus, the key is to control the norm of the perturbation. Many deep results from random matrix theory come into play here \citep{Vu2007SpectralNO,Nadakuditi2012GraphSA,Abbe2020ENTRYWISEEA}.

This nice and simple approach stops working as we step into the sparse regime \citep{Feige2005SpectralTA,CojaOghlan2009GraphPV,Keshavan2009MatrixCF,Decelle2011AsymptoticAO,Krzakala2013SpectralRI}. The main reason is that leading eigenvalues of $A$ are about the order of the square root of the maximum degree. High-degree vertices mess up the desired order of eigenvalues. In particular, for Erdős–Rényi random graphs ($\G (n, d/n)$), we have $\hat{\lambda}_1 = (1+o(1))\sqrt{\log n / \log \log n}$ almost surely \citep{Krivelevich2003TheLE}. Furthermore, the leading eigenvectors are concentrated at these `outliers' with high degree and contain no structural information of the underlying model. 

Take the star graph for example, where we assume that only the first node is connected to $k$ neighbors. It is easy to see that the corresponding adjacency matrix has eigenvalue $\sqrt{k}$ and eigenvector $(\sqrt{k},1,\dots,1)$. Various interesting spectrum-based methods are proposed to overcome this challenge \citep{Mossel2018APO,Massouli2014CommunityDT,Bordenave2015NonbacktrackingSO}. The key idea is to replace adjacency with some combinatorically constructed matrices. However, they typically rely on model statistics and underlying probabilistic assumptions, which leads to the problem of adversarial robustness. For example, they are
non-robust to ’helpful’ perturbations. Namely, if we allow an adversary to perform the following changes on the graph: (1)~adding edges within communities and/or
(2)~removing edges across communities, spectral approaches are going to fail. It is surprising since, intuitively, these changes help to emphasize community structures. 

Meanwhile, semidefinite programming (SDP) sheds light on how we may be able to overcome the limitations of spectral algorithms, which are shown to be robust when SNR is sufficiently large \citep{Moitra2016HowRA}. It is another major line of work on clustering and community detection concerning the performance of SDPs on SBMs. While a clear picture of the unbounded degree case has been figured out by \citep{Abbe2016ExactRI, Hajek2016AchievingEC, Amini2014OnSR, Bandeira2018RandomLM, Agarwal2015MultisectionIT, Perry2017ASP}, the results for sparse networks are more complicated. \citep{Gudon2014CommunityDI} proved a sub-optimal condition, SNR $\geq10^4$, using Grothendieck inequality. Then, with a Lindeberg interpolation process \citep{TTao}, Montanari et al.\ proved that an SDP algorithm as proposed by \citep{montanari2015semidefinite} is nearly optimal for the case of large bounded average degree by transferring analysis of the original SDPs to the analysis of SDPs of Gaussian random matrices.
\begin{theorem}\citep{montanari2015semidefinite}\label{MS}
Assume $G \sim G(n,a/n,b/n)$. If for some $\epsilon > 0$, $\text{SNR}\geq 1+\epsilon$ and $d>d^*(\epsilon)$ then the SDP estimator solves the weak recovery.
\end{theorem}
The necessary degree $d^*$ depends on the choice of $\epsilon$ and goes to infinity as $\epsilon \to 0$. If we fix $d$ and view $\epsilon$ as its function, the condition becomes $\text{SNR} \geq 1 +o_d(1)$. Numerical estimation and non-rigorous statistical mechanism approximation suggest that it is at most $2\%$ sub-optimal. This result seems to be the ceiling of SDP according to the preliminary calculation from \citep{Javanmard2016PhaseTI}. Moreover, they address the irregularity of high-degree nodes by showing SDPs return similar results for Erdős–Rényi random graphs and random regular graphs, which appear to be sensitive only to the average degree. See Section \ref{CSDP} for more discussion on the estimation. Inspired by their work, we propose a natural modification of SDP to incorporate revealed labels in the semi-supervised setting and show that it not only achieves but even crosses, the KS threshold. In turn, our result brings a new perspective to study the (non-)achievability and robustness of (unsupervised) SDPs. 

\subsection{Semi-Supervised Learning}
Within machine learning, there are three basic approaches: supervised learning, unsupervised learning, and the combination of both, semi-supervised learning. The main difference lies in the availability of labeled data. While unsupervised learning (e.g.,\ clustering, association, and dimension reduction) operates without any domain-specific guidance or preexisting knowledge, supervised learning (e.g.,\ classification and regression) relies on all training samples being associated with labels. However, it is often the case where existing knowledge for a problem domain doesn't fit either of these extremes. 

In real-world applications, unlabeled data comes with a much lower cost not requiring expensive human annotation and laboratory experiments. For example, documents crawled from the Web, images obtained from surveillance cameras, and speech collected from the broadcast are relatively more accessible compared to their labels which are required for prediction tasks, such as sentiment orientation, intrusion detection, and phonetic transcript. Motivated by this labeling bottleneck, the semi-supervised approach utilizes both labeled and unlabeled data to perform learning tasks faster, better, and cheaper. Since the 1990s, semi-supervised learning research has enjoyed an explosion of interest with applications like natural language processing \citep{Qiu2019GraphBasedSL, Chen2016StatisticalComputationalTI} and computer vision \citep{Xie2020SelfTrainingWN, Lee2013PseudoLabelT}.

This paper is closely related to the subtopic called constrained clustering, where one has some must-links (i.e.,\ two nodes belong to the same cluster) and cannot-links (i.e.,\ two nodes are in different clusters) as extra information. Although constrained versions of classic algorithms have been studied empirically, such as expectation–maximization \citep{Shental2003ComputingGM}, k-means \citep{Wagstaff2001ConstrainedKC}, and spectral method \citep{Kamvar2003SpectralL}, our methods take different approaches than hard-coding these pairwise constraints into the algorithms and provide theoretically insightful guarantees under SBM.

There also has been some excellent work considering introducing node information into SBMs. One interesting direction is overlaying a Gaussian mixture model with SBM, namely at each node of the graph assuming there is a vector of Gaussian covariates, which are correlated with the community structure. \citep{Yan2016CovariateRC} proposed an SDP with k-means regularization and showed that such node information indeed improves the clustering accuracy, while \citep{Lu2020ContextualSB} formally established the information-theoretic threshold for this model when the
average degree exceeds one. The node information is depicted as noisy observations for all nodes in \citep{Mossel2015LocalAF}. In this setting, random guessing is no longer a meaningful baseline. Hence, the authors refer to the Maximum A Posteriori (MAP) estimator instead and show a local belief propagation algorithm's performance converges to the MAP accuracy in various regimes of the model. They also conjectured that those regimes can be extended to the entire domain of $a > b$ with arbitrary, but non-vanishing, strength of node information. We will see in the next section that our result establishes this conjecture in the sense that with arbitrary, but non-vanishing, knowledge of the labels, we can beat the meaningful baseline for all $a > b$.

All of these models require input on every node, which does not fall within the scope of semi-supervised learning. Whereas, we consider a realistic and intuitive generalization of SBM where a small random fraction of the labels is given accurately. \citep{Kanade2014GlobalAL} studied the same model as ours. They demonstrated that a vanishing fraction of labels improves the power of local algorithms. In particular, when the number of clusters diverges, it helps their local algorithm to go below the conjectured algorithmic threshold of the unsupervised case. Elegantly, they also proved the following result, which is closely related to this paper.
\begin{theorem}\citep{Kanade2014GlobalAL}\label{vanishing}
    When the fraction of revealed node labels is vanishingly small, the (unsupervised) weak recovery problem on the planted bisection model is not solvable if SNR is under the Kesten-Stigum threshold.
\end{theorem}

\section{Our Results}\label{summary}
The main goal of this paper is to answer the long-standing open question regarding semi-supervised learning on probabilistic graph models. We would like to quote the version from \citep{Abbe2017CommunityDA}:
\begin{quotation}
"How do the fundamental limits change in a semi-supervised
setting, i.e.,\ when some of the vertex labels are revealed, exactly or probabilistically?"
\end{quotation}

In the previous section, we discussed deep research related to the clustering/community detection problem on the SBM. Establishing the phase transition phenomenon at the KS threshold is a major focal point. However, such a sharp and intrinsic limit completely disappears when an arbitrarily small fraction of the labels is revealed. This astonishing change is first observed in \citep{PhysRevE.90.052802} where the authors provide non-trivial conjectures based on calculations of the belief propagation approach. 

The theory of semi-supervised clustering contains some fascinating and fundamental algorithmic challenges arising from both the sparse random graph model itself and the semi-supervised learning perspective. To address them rigorously, we first define the semi-supervised SBM so that it captures the essence of realistic semi-supervised learning scenarios and is a natural and simple generalization of unsupervised models. 

\begin{definition}[Semi-supervised planted bisection model]\label{SPBM}
For $n\in \mathbb{N}$, $p,q \in (0,1)$ and $\rho \geq 0$, let $\G(n,p,q,\rho)$ denote the distribution over graphs with $n$ vertices and $n$-dimensional vectors defined as follows. The vertex set is partitioned uniformly at random into two subsets $S_1, S_2$ under the balance constraint $|S_1|=|S_2| = n/2$. Then, conditioned on the partition, two processes are undertaken independently:
\begin{itemize}
    \item Let $E$ denote the edge set of the graph $G$. Edges are included independently with probability defined as follows:
    \begin{equation}
    \prob\left((i,j)\in E | S_1, S_2\right)=
    \begin{cases}
    p & \text{~if~} \{i,j\} \subseteq S_1 \text{~or~} \{i,j\} \subseteq S_2, \\
    q & \text{~if~} i \in S_1, j \in S_2 \text{~or~} i \in S_2, j \in S_1.
    \end{cases}
    \end{equation}
    \item An index set $\mathcal{R}$ of size $m \coloneqq 2\lfloor \rho \cdot \frac{n}{2} \rfloor$ is chosen uniformly at random such that $|\mathcal{R} \cap S_1| = |\mathcal{R} \cap S_2| = m/2$. The revealed labels are given as 
    \begin{equation}
    \Tilde{x}_i=\begin{cases}
    ~1 \quad &i \in \mathcal{R} \cap S_1,\\
    -1 \quad &i \in \mathcal{R} \cap S_2,\\
    ~0 \quad &\text{otherwise.}
    \end{cases} 
    \end{equation}
\end{itemize}
\end{definition}

\begin{remark}
The revealing process is independent of edge formation, i.e.,\ $G \perp \tilde{x}|S_1,S_2$. Moreover, if we set $\rho = 0$ or simply ignore the revealed labels, the model is exactly the unsupervised SBM. In other words, the marginal distribution of the random graph is indeed $\G(n,p,q)$ from Definition \ref{PBM}.
\end{remark}

\begin{remark}
One can also consider revealing uniformly at random over the index set independent of $\G(n,p,q)$ (instead of requiring revealed communities to have the same size), but this modification makes almost no difference in the context of this work. In practice, one can always achieve the balance requirement by either sampling a few more or dropping the uneven part.   
\end{remark}

\begin{remark}
The definition is versatile in the sense that it keeps the unsupervised setting as a special case (and with it all the interesting phase transitions). On the other hand, it can be easily generalized to the multiple and/or asymmetric communities case. 
\end{remark}

Under the semi-supervised setting, we naturally extend community detection problems in a non-trivial way that includes the unsupervised situation as a special case and captures its essence. We will discuss these items in detail when it comes to the corresponding section.

\begin{definition}
    Semi-supervised weak recovery: finding an estimator to perform substantially better than chance \textbf{on the unrevealed vertices}. \\
    Semi-supervised distinguishability: finding a test that, with high probability, distinguishes $\G(n,d/n,d/n,\pmb{\rho})$ from $\G(n,a/n,b/n,\pmb{\rho})$ where $d=\frac{a+b}{2},~a>b$.
\end{definition}

Based on the fact that a $\ln (n)$-neighborhood in $(G,x) \sim \G(n,a/n,b/n)$ asymptotically has the identical distribution as a Galton-Watson tree with Markov process, we propose our first semi-supervised clustering algorithm, called \textit{census method}. Namely, we decide the label estimation of a certain node according to the majority of its revealed neighbors, 
\begin{equation}
\hat{x}_v = \text{sgn}\left( \sum_{i \in \{u \in \mathcal{R}:~ d(u,v)=t\}} x_i \right),
\end{equation}
where $d(u, v)$ is the length of the shortest path connecting $u$ and $v$. We conclude that when $\text{SNR} \leq 1$, the optimal choice of $t$ is indeed 1.
\begin{theorem}\label{thm:census}
    The 1-neighbors census method solves the semi-supervised weak recovery problem with any reveal ratio $\rho>0$ for arbitrary $\text{SNR} >0$.
\end{theorem}

Furthermore, we derive an explicit constant bound for the overlap with the corresponding tail probability. Namely, under $\G(n,a/n,b/n,\rho)$, if we denote \textit{Overlap} on the unrevealed nodes, between 1-neighbors census estimator and the ground truth, as $\Theta$ and let $\delta = \frac{\rho(a-b)}{2\me^{\rho(a+b)}}$, then
$
    \prob\left(\Theta \geq \frac{\delta}{2}\right) \geq 1 - \me^{-\frac{\delta^2 (1-\rho)n}{8}}
$.
A detailed discussion is provided in the next section.

Note that if $\rho \to 0$, semi-supervised weak recovery is equivalent to unsupervised weak recovery. Therefore, Theorem \ref{vanishing} implies that our result is also sharp in the sense of minimum requirement on the fraction of revealed labels.

Although it successfully solves the weak recovery problem in the largest possible regime, some limitations are hindering the census method's utility in practice. Its performance depends on a sufficient amount of revealed labels, hence requiring $n$ to be quite large. Besides, without an unsupervised counterpart, it is not applicable when revealing is unreliable.

To address these challenges, we propose our second semi-supervised clustering algorithm which performs well in practice and covers the unsupervised setting as a special case. As discussed in the previous section, SDPs enjoy many nice properties, among which the monotone-robustness is particularly interesting to us. In the semi-supervised setting, the revealed labels are supposed to enhance the community structure. However, the work from \citep{Moitra2016HowRA} suggests such enhancement may not help with, but to the contrary can hurt the performance of many algorithms, which makes SDP an ideal starting point for us. We define the \textit{Constrained Semidefinite Program} (CSDP) as
\begin{equation}\CSDP(M) = \max_{\substack{X \succeq 0 \\ X_{ii}=1, ~\forall i \in [n]}}\{\la M,X\ra:~ X_{ij} = x_i \cdot x_j, ~\forall i,j \in \mathcal{R}\}\end{equation}
and show that it solves the semi-supervised community detection problem in the form of hypothesis testing. 

\begin{theorem}\label{thm:CSDP}
Let $(G,x_\mathcal{R})\sim \G(n,a/n,b/n,\rho)$ and $A$ be the adjacency matrix associated with $G$. For any $a > b$, there exists $\rho_0 < 1$ such that if $\rho \geq \rho_0$, the CSDP-based test $T(G, x_\mathcal{R};\Delta) = \mathbbm{1}_{\{\CSDP(A - \frac{d}{n} \mathbf{1}\mathbf{1}^\top) \geq n[(a-b)/2 - \Delta]\}}$ will succeed with a high probability for some $\Delta > 0$.
\end{theorem}

\subsection{Proof Techniques}\label{proof tech}
The technical challenges of establishing Theorem \ref{thm:census} root in the fact that the advantage created by revealed labels can be easily blurred out by various approximations of the limit distribution. Instead of the central limit theorem, one needs a Berry–Esseen-type inequality to derive a more quantitative result of the convergence rate. Moreover, since the distribution of each underlying component also depends on $n$, the conventional sample mean formulation does not apply here. We overcome the difficulty above with a direct analysis of non-asymptotic distributions, which leads to a detailed comparison between two binomial variable sequences with constant expectations. 

It is quite surprising that this calculation can be carried out in a rather elegant manner since many other applications of this method are much more technically involved. For example, to establish independence among estimators, one may need to consider the `leave-one-out' trick. But in our case, it comes in a very natural way.

Regarding CSDP, we first show it can be coupled to an SDP with the surrogate input matrices. Moreover, its optimal value lies between two unsupervised SDPs associated with the same random graph model (different parameters). It means that all the analytical results from SDP research can be transferred into the CSDP study. However, we notice that it is common to make assumptions on the average degree $d$ in the relevant literature. It is quite reasonable in the unsupervised setting since the graph topology is a strong indicator of the possibility of weak recovery. E.g.,\ when $d\leq1$, there will not exist a giant component that is of size linear in $n$. 

To establish our result without such extra assumptions, we derive a probabilistic bound on the cut norm of the centered adjacency matrix and then use Grothendieck's inequality to bound the SDP on ERMs from above. This idea follows from \citep{Gudon2014CommunityDI}; we give a slightly different analysis fitting for our purpose. A generalized weak law of large numbers is also derived to address the issue that distributions of the entries change as $n\to \infty$. Then we conclude the proof with a lower bound of the CSDP on SBMs considering a witness that consists of the ground truth of labels.

\subsection{Outline}
The rest of the paper is organized in the following way. In Section \ref{Census method}, we formally derive the census method and prove that it can solve the weak recovery problem throughout the entire parameter domain. In Section \ref{CSDP}, we introduce the constrained SDP and the associated hypothesis test, through which we show that even under the KS threshold (also the information-theoretic threshold), the ERMs and the SBMs become distinguishable in the semi-supervised setting. Section \ref{numexp} includes some numerical simulation results. We end the paper with concluding remarks in Section \ref{conclusion}.

\subsection{Notation}
For any $n\in\mathbb{N}$, we denote the first $n$ integers by $[n] = \{1,2,\dots,n\}$. For a set $S$, its cardinality is denoted by $|S|$. We use lowercase letters for vectors (e.g.,\ $v = (v_1,v_2,\dots,v_n)$) and uppercase letters for matrices (e.g.,\ $M = [M_{ij}]_{i,j\in[n]}$). In particular, for adjacency matrices, we omit their dependency on underlying graphs. Instead of $A_G$, we simply write $A$. $\1_n = (1,\dots,1) \in \mathbb{R}^n$ stands for the all-ones vector, and $I_n$ is the $n\times n$ identity matrix. $\mathbf{e}_i \in \mathbb{R}_n$ represents the $i$'s standard basis vector. For two real-valued matrices $A$ and $B$ with the same dimensions, we define the Frobenius inner product as $\la A, B \ra = \sum_{i,j}A_{ij}\cdot B_{ij} = \Tr (A^\top B)$. Vector inner product is viewed as a special case of $n\times 1$ matrices. Let $\|v\|_p = (\sum_{i=1}^p \|v_i\|^p)^{1/p}$ be the $\ell_p$ norm of vectors with standard extension to $p = \infty$. Let $\|M\|_{p\to q} = \sup_{\|v\|p \leq 1} \|Mv\|_q $ be the $\ell_p$-to-$\ell_q$ operator norm and $\|M\|_{\text{op}} \coloneqq \|M\|_2 \coloneqq \|M\|_{2\to 2}$.
Random graphs induce measures on the product space of label, edge, and revealed node assignments over $n$ vertices. For any $n\in \mathbb{N}$, it is implicitly understood that one such measure is specified with that graph size. The terminology \textit {with high probability} means ‘with probability converging to $1$ as $n \to \infty$’. Also, we follow the conventional Big-Oh notation for asymptotic analysis. $o_p(1)$ stands for convergence to $0$ in probability.

\section{Census Method}\label{Census method}
Analysis of the model from Definition \ref{PBM} is a challenging task since conditioned on the graph, it is neither an Ising model nor a Markov random field. This is mainly due to the following facts: (1) The balance requirement puts a global condition on the size of each cluster; (2) Even if conditioned on sizes, there is a slight repulsion between unconnected nodes. Namely, if two nodes do not form an edge, the probability of them being in the same community is different from the probability of them being in opposite communities. 

Recent years have witnessed a series of excellent contributions to the study of phase transitions in the sparse regime. Our census method for semi-supervised clustering is mainly inspired by the natural connection between community detection on SBMs and reconstruction problems on trees, which was formally established by \citep{MosselNS15}. Intuitively, for a vertex $v$ in $\G(n,a/n,b/n)$, it is not likely that a node from its small neighborhood has an edge leading back to $v$. Therefore, the neighborhood looks like a randomly labeled tree with high probability. Furthermore, the labeling on the vertices behaves like broadcasting a bit from the root of
a tree down to its leaves (see the survey \citep{Mossel2001SurveyIF} for a detailed discussion).

In this section, we will first look into the census method of $t$-neighbors, i.e.,\ deciding the label of a node by the majority on its neighbors at depth $t$. After defining the general framework, we will show that when $\text{SNR}\leq 1$, $1$-neighbors voting is optimal in terms of recovering the cluster structure via informal calculation. Then, we rigorously prove that census on $1$-neighbors solves the semi-supervised weak recovery problem for any $\text{SNR}>0$ with an arbitrarily small fraction of the labels revealed. 

\subsection{Majority of t-Neighbors}
Let $(G,x)$ obey the planted bisection model $\G(n,a/n,b/n)$. We denote the set of all vertices by $V(G)$. For a fixed vertex $v$ and $t\in\mathbb{N}$, let $N_t(v)$ denote the number of vertices that are $t$ edges away from $v$. $\Delta_t(v)$ is defined as the difference between the numbers of $t$-neighbors in each community. Namely,

\begin{align}
    N_t(v) &= |K_t(v)| \\
    \Delta_t(v) &= \sum_{u \in K_t(v)} x_{u} 
\end{align}
where $K_t(v)\coloneqq\{u \in V(G):~ d(u,v)=t\}$ denotes the $t$-neighbors of $v$.

If one assumes that the subgraph of $G$ induced by the vertices within $t$ edges of $v$ is a tree, the expected value of $N_t(v)$ is approximately $[(a+b)/2]^t$, and the expected value of $x_v \cdot \Delta_t(v)$, i.e.,\ the expected number of these vertices in the same community as $v$ minus the expected number of these vertices in the other community, 
is approximately $[(a-b)/2]^t$. So, if one can somehow independently determine
which community a vertex is in with an accuracy of $1/2 + \alpha$ for some $\alpha > 0$, one will be able to predict the label of each vertex with an accuracy of roughly $1/2 + [(a-b)^2/(2(a+b))]^{t/2}\cdot\alpha$, by guessing it as the majority of $v$'s $t$-neighbors. Under the unsupervised learning setting, one can get a small advantage, $\alpha \sim \Theta(1/\sqrt{n})$, by randomly initializing labels. It is guaranteed by the central limit theorem that such a fraction exists in either an agreement or disagreement form.

\begin{figure}[ht]
    \centering\includegraphics[width=0.8\textwidth]{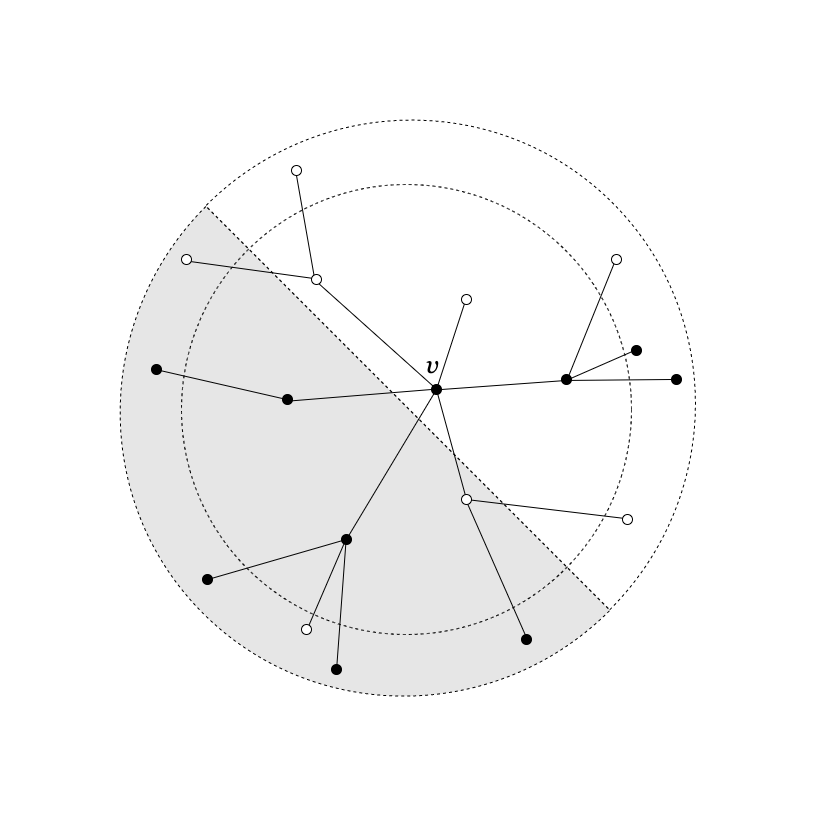}
    \caption{Neighborhood of node $v$ with a tree structure. The ground truth of clusters is coded in black and white. The shaded area indicates those nodes randomly guessed to be in the same community or the opposite community as $v$. The annulus represents the collection of its $t$-neighbors.}
    \label{fig:census}
\end{figure}

To amplify this lucky guess, we need $t$ to be sufficiently large so that $[(a-b)^2/(2(a+b))]^{t/2} > \sqrt{n}$, which implies $[(a+b)/2]^t>n$. Note that $d = (a+b)/2$ is the average degree. This means before the signal is strong enough for our purpose, not only our tree approximation will break down, but vertices will also be exhausted. However,
if we have access to some of the true labels, i.e.,\ in the semi-supervised setting, we can leverage the tree structure to get a non-vanishing advantage over random guessing. 

Let $A$ be the adjacency matrix associated with $G$. Consider the random variables $Y_u$ representing votes of directly connected neighbors,
\begin{equation}
    Y_u = 
    \begin{cases}
    x_{u} &\text{if~} A_{uv} = 1, \\
    0 &\text{otherwise}.
    \end{cases}
\end{equation}
We have
\begin{align}
    N_1(v) &= \sum_{u \in V(G)} |Y_u|, \\
    \Delta_1(v) &= \sum_{u \in V(G)} Y_u.
\end{align}
By definition of the planted bisection model,
\begin{equation}
    \prob(Y_u=1 | x_v=1) = \frac{\prob(Y_u=1, x_v=1)}{\prob(x_v=1)} \approx \frac{a}{2n}.
\end{equation}
Similarly,
\begin{equation}
    \prob(Y_u=-1 | x_v=1) \approx \frac{b}{2n}.
\end{equation}
It is not exact due to the balanced community constraint. But when $n$ is large, such an effect is negligible. Furthermore, if we consider the definition of the planted bisection model without balance constraint, the equation will be exact. 

Without loss of generality, we only consider the case where $x_v = 1$ and omit the condition on it. We have

\begin{equation}
\Delta_1(v)=\sum_{u \in V(G)} Y_u \qquad\text{with~~} Y_u =
\begin{cases}
1 & \text{w.p.~} \frac{a}{2n}\\
-1 & \text{w.p.~} \frac{b}{2n}\\
0 & \text{w.p.~} 1-\frac{a+b}{2n}
\end{cases}
\end{equation}
where the $Y_u$'s are independent. Note that $\E (Y_u) = \frac{a-b}{2n}$ and $\E (Y_u^2) = \frac{a+b}{2n}$.

Recall that $\rho \in [0,1]$ is the ratio of revealed labels. For the sake of simplicity, we assume the total number of revealed vertices $m = \rho n \in 2\mathbb{N}$ to be an even integer. The revealed vertices are chosen arbitrarily, denoted as $\mathcal{R} \coloneqq \{u_{n-m+1},u_{n-m+2},\dots,u_{n}\}$. The model also provides that the number of revealed vertices in each community is $\frac{\rho n}{2}$. Then the majority of revealed vertices among 1-neighborhood of $v$ can be written as

\begin{equation}
    \Tilde{\Delta}_1(v)=\sum_{u \in \mathcal{R}} Y_u.
\end{equation}

Therefore,
\begin{align}
    &\E (\Tilde{\Delta}_1(v)) = \sum_{u\in \mathcal{R}} \E(Y_u)
    = \rho \frac{a-b}{2},\\
    &\V (\Tilde{\Delta}_1(v)) = \sum_{u\in \mathcal{R}} \V(Y_u)
    = \rho \frac{a+b}{2} + o(1).
\end{align}

\subsection{Locally Tree-Like Structure}
Proceeding to the $t$-neighbors, we need to first understand the structure of a small neighborhood in the SBM. The neighborhoods in a sparse network locally have no loops. So they have a nice tree-like structure. Moreover, the labels also obey some random broadcasting processes on trees. 

A broadcasting process transmits the information from the root of a tree to all the nodes. At each level, nodes inherit the information from its parent. Meanwhile, errors could happen with a certain amount of probability. Usually, the edges are assumed to be included according to the same rule and work independently. It was first considered in genetics \citep{CAVENDER1978271} since it perfectly describes the propagation of a gene from ancestor to descendants. It can also be interpreted as a communication network that passes out the information from the root. Such processes were intensively studied in information theory and statistical physics \citep{10.2307/2959462, Higuchi1977RemarksOT, Bleher1995OnTP}. In particular, we are interested in the following Markov process since it can be identified with the labeling process of a small neighborhood in SBM.

\begin{definition}[Galton–Watson tree with Markov process]
    Let $T$ be an infinite rooted tree with root $v$. Given a number $0\leq\epsilon<1$ and the offspring rate $d>0$, we define a random labeling $\tau \in \{1,-1\}^T$. First, draw $\tau_v$ uniformly in $\{1,-1\}$. Then, recursively construct the labeling as follows:  
    \begin{itemize}
        \item Generate children of each parent node according to a Poisson distribution with mean $d$.
        \item Conditionally independently given $\tau_v$, for every child $u$ of $v$, set $\tau_u=\tau_v$ with probability $1-\epsilon$ and $\tau_u=-\tau_v$ otherwise.
    \end{itemize} 
\end{definition}

The following lemma shows that a $\ln (n)$-neighborhood in $(G,x)$ looks like a Galton-Watson tree with Markov process. For any $v \in G$, let $G_R$ be the induced subgraph on $\{u \in G : d(u, v) \leq R\}$.

\begin{lemma}\citep{MosselNS15}
    Let $R = R(n) = \frac{\ln n}{10\ln(2(a+b))}$. There exists a coupling between $(G, x)$ and $(T, \tau)$ such that $(G_R, x_{G_R} )$ = $(T_R, \tau_{T_R})$ a.a.s.
\end{lemma}

Hence, for fixed $t \in \mathbb{N}$, $t \leq R$, and any $v \not\in \mathcal{R}$, we can denote the label of a vertex in $v$'s $t$-neighborhood as $Y_i^{(t)} \coloneqq \Pi_{k=1}^t \prescript{k}{}{Y_u}$, where $\{\prescript{k}{}{Y_u}\}_{k=1}^t$ are independent copies of $Y_u$.  Then we have $\E (Y_i^{(t)}) = (\frac{a-b}{2n})^t$ and $\E ((Y_i^{(t)})^2) = (\frac{a+b}{2n})^t$. Moreover, $\{Y_i^{(t)}\}$'s are essentially independent. Therefore, the census of $v$'s revealed $t$-neighbors can be written as 

\begin{equation}
    \Tilde{\Delta}_t(v)=\sum_{i \in [\rho \cdot n^t]} Y_i^{(t)} \quad \textit{(a.a.s)}.
\end{equation}

The central limit theorem suggests
\begin{equation}
    \Tilde{\Delta}_t(v) \to \mathcal{N}(\rho(\frac{a-b}{2})^t, \rho(\frac{a+b}{2})^t), \quad \text{as~} n \to \infty .
\end{equation}

Hence, 
\begin{align}
    \prob(\Tilde{\Delta}_t(v) >0 | x_v=1) &= \frac{1}{2}\left[1 + \text{erf}\left(\frac{\rho[(a-b)/2]^t}{\sqrt{\rho[(a+b)/2]^t}\sqrt{2}}\right)\right] + o(1)\\
    &= \frac{1}{2}+ \frac{1}{2}\text{erf}\left(\sqrt{\frac{\rho~\text{SNR}^t}{2}}\right) + o(1)
\end{align}
where $\text{erf}(x) = \frac{2}{\sqrt{\pi}} \int_0^x \exp(-t^2) \dif{t} $ is the Gauss error function.

So one can see that once SNR is less than or equal to $1$, it is not beneficial to look into $t$-neighbors. The optimal choice of $t$ is $1$ in this situation. Since we also know that weak recovery is solvable when $\text{SNR}>1$, it makes the majority of $1$-neighbors particularly interesting.

Suppose $\text{SNR}\leq 1$ and include the symmetric part of $x_v=-1$, we have 
\begin{equation}
    \prob(\text{sgn}(\Tilde{\Delta}_1(v)) = x_v)> \frac{1}{2} + \frac{1}{3}\sqrt{\rho~\text{SNR}}.
\end{equation}

Consider the estimator of unrevealed labels

\begin{equation}\hat{x}_{\mathcal{R}^\complement}\coloneqq \text{sgn}\left([\Tilde{\Delta}_1(u_1), \Tilde{\Delta}_1(u_2), \dots, \Tilde{\Delta}_1(u_{n-m})]^\top\right)\end{equation}
and the ground truth $x_{\mathcal{R}^\complement}=[x_{u_1},x_{u_2},\dots,x_{u_{n-m}}]^\top$. Recall that

\begin{equation}
    \text{Overlap}(x_{\mathcal{R}^\complement},\hat{x}_{\mathcal{R}^\complement})=\frac{1}{n-m}| \langle x_{\mathcal{R}^\complement},\hat{x}_{\mathcal{R}^\complement} \rangle |.
\end{equation}

We can conclude that
\begin{align}
    \E [\text{Overlap}(x_{\mathcal{R}^\complement},\hat{x}_{\mathcal{R}^\complement})] &= \E\left[\frac{1}{n-m} \left|\sum_{i\in[n-m]}\text{sgn}(\Tilde{\Delta}_1(u_i))x_{u_i}\right|\right]\\
    &\geq \frac{1}{n-m}\left|\sum_{i\in[n-m]} \E\left[\text{sgn}(\Tilde{\Delta}_1(u_i))x_{u_i}\right]\right|\\
    &> \frac{2}{3}\sqrt{\rho~\text{SNR}}.
\end{align}

\begin{figure}[htbp]
    \centering\includegraphics[width=\textwidth]{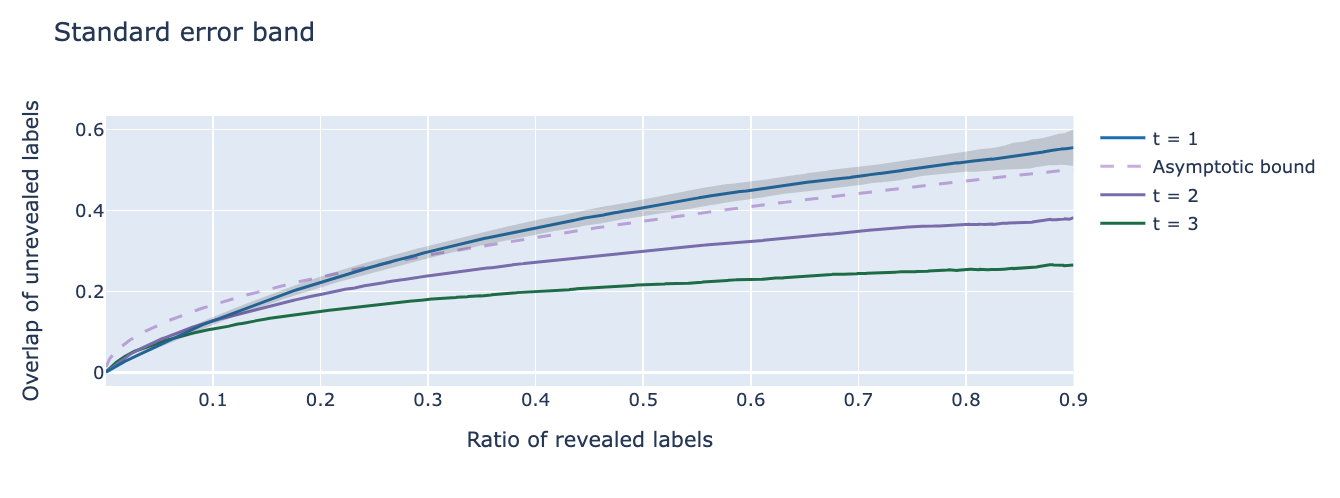}
    \caption{The simulation result of $\G(3000, 5/3000, 2/3000)$, $\text{SNR}\approx0.64$. Solid curves stand for the average overlaps of the t-neighbors census method ($t$ = 1, 2, and 3) on 60 independent realizations of the random graph. The shaded area represents the standard error band of the 1-neighbors census. The dashed curve stands for the asymptotic lower bound we conclude from our calculation.}
    \label{fig:rho_overlap}
\end{figure}

The expected overlap is not vanishing which suggests that weak recovery is solvable for any SNR. But it is technically impractical to rigorously describe the limit distribution of our census estimator without blurring this edge out. From Figure \ref{fig:rho_overlap}, we can see that our calculation is close to the expectation. But the convergence rate depends on $\rho$. In particular, when both SNR and $\rho$ are small, the asymptotic behavior of our algorithm remains unclear. Hence, we go through a direct analysis to establish the desired result.

\subsection{Majority of 1-Neighbors}

Since the algorithm is invariant under index reordering, without loss of generality, we let the adjacency matrix $A$ be a symmetric matrix with diagonal entries $A_{ii}=0,~ i=1,2, \dots, n$. For $1 \leq i < j \leq n$, $\{A_{ij}\}$'s are independent,

\begin{align}
    &A_{ij} \sim \text{Bernoulli}\left(\frac{a}{n}\right) &\quad \left(i\leq \frac{n}{2} ~\text{and}~ j \leq \frac{n}{2} \right) &~\text{or}~ \left(i\geq \frac{n}{2} ~\text{and}~ j \geq \frac{n}{2} \right),\\
    &A_{ij} \sim \text{Bernoulli}\left(\frac{b}{n}\right) &\quad i\geq \frac{n}{2} ~&\text{and}~ j \leq \frac{n}{2}.
\end{align}

The true label $x$ and revealed label $\Tilde{x}$ are, respectively,

\begin{equation}
    x_i=\begin{cases}
    1 \quad &i=1,2,\dots,\frac{n}{2},\\
    -1 \quad &i=\frac{n}{2},\frac{n}{2}+1,\dots,n;
    \end{cases}
    \quad
    \Tilde{x}_i=\begin{cases}
    1 \quad &i=1,2,\dots,\frac{m}{2},\\
    -1 \quad &i=\frac{n}{2},\frac{n}{2}+1,\dots,\frac{n+m}{2},\\
    0 \quad &\text{otherwise.}
    \end{cases} 
\end{equation}

For an unrevealed vertex, we consider the majority of its 1-neighbors,
\begin{equation}
    \Tilde{\Delta}_1(i) = \langle A[i,:],\Tilde{x} \rangle = \sum_{j:\Tilde{x}_j \neq 0} A_{ij}\Tilde{x}_j = \sum_{j:\Tilde{x}_j \neq 0} A_{ji}\Tilde{x}_j.
\end{equation}
Therefore, $\{\Tilde{\Delta}_1(i)\}$'s are independent for all $i:\Tilde{x}_i = 0$ since they have no common term. Note that this is not the case for all $i \in [n]$. However, we only need to predict the unrevealed labels, hence the independence holds.

The estimator given by majority voting of 1-neighbors is
\begin{equation}
    \hat{x}_i = \begin{cases}
    \Tilde{x}_i \quad &\text{if}~\Tilde{x}_i \neq 0,\\
    \text{sgn}^*(\Tilde{\Delta}_1(i)) \quad &\text{if}~\Tilde{x}_i = 0.
    \end{cases}
\end{equation}
We toss a fair coin when $\Tilde{\Delta}_1(i)=0$ to break the tie, i.e.,\ \begin{equation}\prob(\text{sgn}^*(\Tilde{\Delta}_1(i))=1|\Tilde{\Delta}_1(i)=0) = \prob(\text{sgn}^*(\Tilde{\Delta}_1(i))=-1|\Tilde{\Delta}_1(i)=0) = \frac{1}{2}.\end{equation}
Formally, for a finite sequence of random variables $W_i$, 
\begin{equation}
    \text{sgn}^*(W_i) = \text{sgn}(W_i) + \mathbbm{1}_{\{W_i = 0\}} (2B_i - 1)
\end{equation} 
where $\{B_i\}$'s are i.i.d.~Bernoulli(1/2) random variables, independent of all other random objects. Note that it is only introduced for analysis purposes and is equivalent to the conventional sign function in practice. 

Suppose $(G,x)$ is an Erdős–Rényi random graph with revealed label $\Tilde{x}$, any estimator can only have a vanishing correlation with the true label among the unrevealed vertices. So the semi-supervised weak recovery problem on SBM requires finding an estimator such that the correlation restricted on the unrevealed part is non-vanishing. Formally, we want to show that

 \begin{equation}\prob\left(\text{Overlap}(x|_{\Tilde{x}_i =0},\hat{x}|_{\Tilde{x}_i =0}) \geq \Omega(1) \right)=1-o(1).\end{equation}

As discussed in Section \ref{proof tech}, we start with a critical result scrutinizing binomial variable sequences. It gives us an edge over direct analysis via a Berry–Esseen-type inequality, which usually assumes the distribution of individual random variables in the sequence independent of $n$.
\begin{lemma}\label{constant gap} 
Let $X$ and $Y$ be two independent binomial random variables with $X \sim \text{Binomial}(n,\frac{a}{n})$ and $Y \sim \text{Binomial}(n,\frac{b}{n})$, $a>b$. Denote $\delta = \delta(a,b) \coloneqq \frac{a-b}{2\exp{(a+b)}}$. Then, for sufficiently large $n$,
\begin{equation}\prob(X>Y) - \prob(X<Y) \geq \delta .\end{equation}
\end{lemma}

\begin{remark}
By symmetry, we always have $\prob(X>Y) - \prob(X<Y) > 0$. This lemma guarantees the difference will not vanish as $n\to \infty$.
\end{remark}

\begin{proof}
By the law of total probability and independence, we have
\begin{align}
    \prob(X>Y)&=\sum_{x=1}^n \prob(Y<x)\prob(X=x)\\
    &=\sum_{x=1}^n\sum_{y=0}^{x-1} \prob(Y=y)\prob(X=x)\\
    &=\sum_{x=1}^n\sum_{y=0}^{x-1} \left[ \binom{n}{x}\left(\frac{a}{n}\right)^x \left(1-\frac{a}{n}\right)^{n-x} \binom{n}{y}\left(\frac{b}{n}\right)^y \left(1-\frac{b}{n}\right)^{n-y} \right].
\end{align}
Let $\Delta \coloneqq \prob(X>Y) - \prob(X<Y)$, then
\begin{multline*}
    \Delta=\sum_{x=1}^n \binom{n}{x} \left(\frac{a}{n}\right)^x\left(1-\frac{a}{n}\right)^{n-x}\left(\frac{b}{n}\right)^x\left(1-\frac{b}{n}\right)^{n-x}\\
    \Bigg\{\sum_{y=0}^{x-1} \binom{n}{y} \left[ \left(\frac{b}{n}\right)^{y-x}\left(1-\frac{b}{n}\right)^{x-y} - \left(\frac{a}{n}\right)^{y-x} \left(1-\frac{a}{n}\right)^{x-y} \right]\Bigg\}
\end{multline*}
\begin{multline*}
    \phantom{\Delta}=\sum_{x=1}^n \binom{n}{x} \left(\frac{ab}{n}\right)^x \left(1-\frac{a+b}{n}+\frac{ab}{n^2}\right)^{n-x}\\
     \Bigg\{\sum_{y=0}^{x-1} \binom{n}{y}\frac{1}{n^y} \left[ \left(\frac{1}{b}-\frac{1}{n}\right)^{x-y} - \left(\frac{1}{a}-\frac{1}{n}\right)^{x-y} \right]\Bigg\}.
\end{multline*}

Let $f(x)= \alpha^x - \beta^x,~ \alpha > \beta >0$. Since $f'(x) = \alpha^x\ln{\alpha} - \beta^x\ln{\beta} >0$, we have $f(m)\geq f(1) = \alpha - \beta,~ \forall m\in \mathbb{N}$. So $\left(\frac{1}{b}-\frac{1}{n}\right)^{x-y} - \left(\frac{1}{a}-\frac{1}{n}\right)^{x-y}\geq \frac{a-b}{ab}$.

Also notice that $\binom{n}{m}= \prod_{i=0}^{m-1}\frac{n-i}{m-i} \geq \left(\frac{n}{m}\right)^m,~ \forall 1\leq m\leq n$. We have
\begin{align}
    \Delta &\geq \sum_{x=1}^n \left(\frac{ab}{x}\right)^x \left(1-\frac{a+b}{n}\right)^{n-x} \left( \sum_{y=0}^{x-1} \frac{1}{y^y}\cdot\frac{a-b}{ab} \right)\\
    &\geq (a-b) \left(1-\frac{a+b}{n}\right)^n\\
    &\geq \frac{a-b}{2\exp{(a+b)}} \qquad (\text{for sufficiently large $n$})
\end{align}
where we follow the convention that $0^0=1$.
\end{proof}

Then, we can simply resort to a classical concentration inequality to bound the overlap.

\begin{lemma}[Chernoff–Hoeffding theorem \citep{Chernoff1952AMO}] \label{Chernoff-Hoeffding}
Suppose $X_1, \dots, X_n$ are i.i.d.~random variables, taking values in $\{0, 1\}$. Let $p = \E(X)$ and $\epsilon > 0$. Then
    \begin{equation}
        \prob\left(\frac{1}{n}\sum X_i \leq p-\epsilon \right) \leq \left( \left(\frac{p}{p-\epsilon}\right)^{p-\epsilon} \left(\frac{1-p}{1-p+\epsilon}\right)^{1-p+\epsilon} \right) ^n =\me^{-D(p-\epsilon\|p)n}
    \end{equation}
    where $D(x\|y)=x\ln\frac{x}{y} + (1-x)\ln(\frac{1-x}{1-y})$ is the Kullback–Leibler-divergence between Bernoulli distributed random variables with parameters $x$ and $y$.
\end{lemma}

We now convert the KL divergence to the total variation distance, which is easier to work with. Let $P_1$ and $P_2$ be two probability measures defined on the same sample space $\Omega$ and sigma-algebra $\mathcal{F}$. The total variation distance between them is defined as $d_{TV}(P_1,P_2) = \sup_{E \in \mathcal{F}}|P_1(E) - P_2(E)|$. Moreover, in the discrete case, we have the following identity $d_{TV}(P_1,P_2) = \frac{1}{2}\|P_1-P_2\|_1 = \sum_{\omega\in \Omega} \frac{1}{2}\|P_1(\omega) - P_2(\omega)\|$. It is related to the KL divergence through Pinsker's inequality (see, e.g.,\ \citep{Tsybakov2009IntroductionTN}, Chapter 3). For completeness, we include an elementary proof of the Bernoulli special case that is sufficient for our usage later.
\begin{lemma} \label{TV-KL}
    Let $P_1$ and $P_2$ be two Bernoulli distributions, where $P_1(1) =x$ and $P_2(1) =y$. We have 
    \begin{equation}2(d_{TV}(P_1,P_2))^2 \leq D(x\|y).\end{equation}
\end{lemma}

\begin{proof}
We can manipulate both sides of the inequality as
    \begin{equation} 
        D(x\|y) = x\ln\frac{x}{y} + (1-x)\ln(\frac{1-x}{1-y}),
    \end{equation}

    \begin{equation}
        2(d_{TV}(P_1,P_2))^2 = \frac{1}{2} \|P_1 - P_2\|_1^2 = 2(x-y)^2.
    \end{equation}
We denote $f(x,y) = x \ln \frac{x}{y} + (1-x)\ln \frac{1-x}{1-y} - 2(x-y)^2$. Therefore,
\begin{equation}
    \frac{\partial f}{\partial y} = (x-y)[4-\frac{1}{y(1-y)}]
\end{equation}
Notice that since $0 \leq y \leq 1$, $y(1-y) \leq \frac{1}{4}$. So $4-\frac{1}{y(1-y)}$ is always negative. Thus, for fixed $x$, $f(x,y) \geq f(x,x) = 0 ,~ \forall y$. Hence, \begin{equation}D(x\|y) - 2(d_{TV}(P_1,P_2))^2 \geq 0.\end{equation}
\end{proof}

Now we prove the main result for the census method.

\begin{proof}[Proof of Theorem \ref{thm:census}]
    Recall that for any $i$ such that $\Tilde{x}_i = 0$, our estimator is defined as $\hat{x}_i = \text{sgn}^*(\Tilde{\Delta}_1(i))$ and
    \begin{equation}\Tilde{\Delta}_1(i) = \sum_{j:\Tilde{x}_j \neq 0} A_{ij}\Tilde{x}_j = \left(\sum_{\rho \frac{n}{2} <j \leq \frac{n}{2}} A_{ij} \right) - \left(\sum_{(1+\rho) \frac{n}{2} <j \leq n} A_{ij} \right).\end{equation}
    
    It is indeed the difference between two independent binomial variables with parameters $(\rho n, \frac{\rho a}{\rho n})$ and $(\rho n, \frac{\rho b}{\rho n})$. By Lemma \ref{constant gap}, we have
    \begin{equation}
        \prob(\text{sgn}(\Tilde{\Delta}_1(i)) = x_i) - \prob(\text{sgn}(\Tilde{\Delta}_1(i)) = - x_i) \geq \delta =  \frac{\rho(a-b)}{2\me^{\rho(a+b)}}
    \end{equation}
     for sufficiently large $n$. Also, notice that
    \begin{equation}  
        \prob(\text{sgn}(\Tilde{\Delta}_1(i)) = - x_i) = 1 - \prob(\text{sgn}(\Tilde{\Delta}_1(i)) = x_i) - \prob(\Tilde{\Delta}_1(i)=0).
    \end{equation}
    Therefore,
    \begin{equation}  
        \prob(\text{sgn}(\Tilde{\Delta}_1(i)) =  x_i) \geq \frac{1+\delta}{2} -  \frac{1}{2}\prob(\Tilde{\Delta}_1(i)=0).
    \end{equation}
    Then, by the law of total probability, we have
    \begin{align}
        \prob(\hat{x}_i = x_i)  &= \prob(\text{sgn}^*(\Tilde{\Delta}_1(i)) = x_i)\\ 
        &= \prob(\text{sgn}(\Tilde{\Delta}_1(i)) = x_i) + \frac{1}{2}\prob(\Tilde{\Delta}_1(i)=0)\\
        &\geq \frac{1}{2} + \frac{\delta}{2}. 
    \end{align}
    
    Since the $\{\hat{x}_i\}$'s are independent for all unrevealed vertices as $\{\Tilde{\Delta}_1(i)\}$'s and $\E \left[\frac{\hat{x}_ix_i+1}{2}\right] = \prob(\hat{x}_i = x_i)$, Lemma \ref{Chernoff-Hoeffding} and Lemma \ref{TV-KL} give us that
    \begin{equation}
        \prob \left(\frac{1}{(1-\rho)n}\sum_{i:\Tilde{x}_i = 0}\frac{\hat{x}_ix_i+1}{2}\leq \frac{1}{2} + \frac{\delta}{2} - \epsilon \right) \leq \me^{-2\epsilon^2 (1-\rho)n}. 
    \end{equation}
Taking $\epsilon = \frac{\delta}{4}$, we have 
\begin{equation}
    \prob\left(\text{Overlap}(x|_{\Tilde{x}_i =0},\hat{x}|_{\Tilde{x}_i =0}) \geq \frac{\delta}{2}\right) \geq 1 - \me^{-\frac{\delta^2 (1-\rho)n}{8}}.
\end{equation}
As long as $a>b$, we have $\delta > 0$, which concludes the proof. 
\end{proof}

\begin{corollary}\label{distinguishable}
The semi-supervised SBM and ERM are not mutually contiguous for any given $a>b\geq 0$ and $\rho>0$.
\end{corollary}

\begin{proof}
Let $P_n^{(0)} = \G(n, \frac{a+b}{2n}, \frac{a+b}{2n}, \rho)$ and $P_n^{(1)}$ = $\G(n, \frac{a}{n}, \frac{b}{n}, \rho)$. Then consider the same constant $\delta > 0$ from the proof of Theorem \ref{thm:census} and denote the event sequence $E_n = \{\text{Overlap}(x|_{\Tilde{x}_i =0},\hat{x}|_{\Tilde{x}_i =0}) \geq \frac{\delta}{2}\}$ where $\hat{x}$ is our semi-supervised census estimator. We have
    \begin{align}
        P_n^{(0)}(E_n) \rightarrow 0 \quad &\text{(Law of large number)},\\
        P_n^{(1)}(E_n) \nrightarrow 0 \quad &\text{(Bounded from below)}.
    \end{align}
\end{proof}

Recall that distinguishability and weak recovery are equivalent in the sense that they share the same phase transition threshold. In the semi-supervised setting, it is straightforward that the weak recovery implies distinguishability. So, Theorem \ref{thm:census} suggests that in this case, SBM and ERM are always distinguishable, which is also equivalent to Corollary \ref{distinguishable}. Hence, the two problems share the same solvable region as well.

\section{Semi-Supervised SDP}\label{CSDP}
We have seen that the census method solves the semi-supervised community detection problem. However, the algorithm is desirable in practice only when the amount of revealed labels is sufficient to support a reasonable performance. In other words, it has no unsupervised `fallback' built-in. Meanwhile, SDPs enjoy nice properties like optimality and robustness as mentioned earlier. It is also well known that approximate information about the extremal cuts of 
graphs can be obtained by computing the optimizer for the SDP of their adjacency matrix, for example \citep{Goemans1995ImprovedAA}. From both a practical and a mathematical point of view, we are interested in developing an SDP-based semi-supervised clustering approach, through which we shall be able to see the models, algorithms, and phase transitions with a fresh perspective. 

In this section, we will focus on the hypothesis testing formulation of the community detection problem. We have discussed the equivalency between it and the non-vanishing overlap formulation under the unsupervised setting. In the semi-supervised scenario, it is still an interesting question to ask whether there exists a test that can distinguish SBMs from ERMs. Here we understand ERM as the special case of SBM with $a = b$. It also has ground truth of labels, which is uniformly random under the balance constraint. Given that they are originally contiguous when $\text{SNR} \leq 1$, we want to show that revealed labels together with random graphs can separate them. 

\subsection{SDP for Community Detection}
Under the Planted Bisection Model specified in Definition \ref{PBM}, the MAP estimator is equivalent to the Maximum Likelihood estimator, which is given by min-bisection, i.e.,\ a balanced partition with the least number of crossing edges. Formally, it can be written the following optimization problem,
\begin{equation}
    \max_{\substack{x\in \{1,-1\}^n \\ x^\top \1 = 0}} x^\top A x.
\end{equation}
    
By lifting the variable $X\coloneqq xx^\top$, we can rewrite it as

\begin{equation}
    \hat{X}_\text{MAP}(G)=\argmax_{\substack{X \succeq 0 \\ X_{ii}=1, ~\forall i \in [n] \\ \text{rank} (X) = 1 \\ X \mathbf{1} = 0}} \langle A,X \rangle.
\end{equation}

Although min-bisection of $G$ is optimal (in the MAP sense) for exact recovery, finding it is NP-hard. Various relaxations have been proposed for the MAP estimator. Since the rank constraint  makes the optimization difficult, we can remove it to make the problem convex. One can also get rid of the balance constraint by centralizing the adjacency matrix, $\tilde{A} \coloneqq A - \frac{d}{n} \mathbf{1}\mathbf{1}^\top$ with $d=(a+b)/2$ the average degree. This can also be justified using Lagrangian multipliers. 
The resulting semidefinite relaxation is given by
\begin{equation}\label{eqn:SDP}
\hat{X}_{\text{SDP}}(G)=\argmax_{\substack{X\succeq 0\\ X_{ii}=1, ~\forall i\in [n]}} \langle \tilde{A}, X \rangle.
\end{equation} 
    
The feasible region $\{X\in\mathbb{R}^{n \times n}: X\succeq0, X_{ii}=1 ~\forall i \in [n] \}$ is indeed the space of correlation matrices, which defines a subset of the unit hypercube and is also called the \textit{elliptope}. Although it is derived from the relaxation of MAP, one can define the SDP for general symmetric matrices as

\begin{equation}\label{eqn:SDP_}
    \SDP(M_{n\times n}) = \max\{\la M,X\ra: X \in \text{elliptope}_n\}.
\end{equation}

\begin{proposition}
For any $n \times n$ symmetric matrix $M$, if we denote its leading eigenvalue as $\lambda_1$, then 
\begin{equation}\frac{1}{n}\SDP(M)\leq \lambda_1.\end{equation}
\end{proposition}
\begin{proof}
For any feasible $X \succeq 0$ and $X_{ii} = 1$, we have $\Tr(X) = n$, moreover,
\begin{equation}
\la X,M = U \Lambda U^\tp \ra = \Tr(U^\tp X U \Lambda) = \la Y \coloneqq U^\tp X U, \Lambda \ra = \sum Y_{ii}\lambda_i \leq n \lambda_1.
\end{equation}
The last inequality follows from $\Tr(Y) = n$ and $Y \succeq 0$ so that $Y_{ii}\geq0$.
\end{proof}

This proposition relates SDPs to spectra of the underlying matrices, which suffer from those high-degree nodes as we mentioned in the introduction. In contrast, SDPs behave similarly on SBMs and random regular graphs. The optimal values of the SDPs for both are approximately $2n\sqrt{d}$, see \citep{montanari2015semidefinite}. Random regular graphs obey the uniform distribution over graphs with $n$ vertices and uniform degree $d$, which provide a simple example to illustrate the regularity property of SDPs. We cite an intermediate result from the original proof as Lemma \ref{lemma:SDP}.

An important way to understand SDPs is by considering the Burer-Monteiro factorization of $X$, which characterizes the constraints. We have 
\begin{equation}\label{BMF}
X=\Sigma \Sigma^\top
\end{equation}
where $\Sigma=(\sigma_1,\sigma_2,\dots, \sigma_n)^\top$ and $\|\sigma_i\|_2=1, ~\forall i \in [n]$. Therefore, the $i$-th node of the graph is associated with the vector $\sigma_i$ that lies on the unit sphere. $X_{ij}=\langle \sigma_i, \sigma_j \rangle$ can be interpreted as the affinity metric between nodes $i$ and $j$. SDP maximizes the likelihood score of this affinity matrix concerning the given centralized adjacency matrix. The optimizer $X^*$ is a better representation of the structure information than the vanilla adjacency matrix. Then we can identify the labels by simply running the k-means method on it or compute the eigenvector corresponding to the largest eigenvalue. 
    
\subsection{Constrained SDP and Semi-Supervised Testing} \label{CSDP def}
In this section, we will introduce our SDP modification and prove that it solves the semi-supervised community detection problem with the hypothesis testing formulation. Let $x$ denote labels of G(n, $\frac{a}{n}$, $\frac{b}{n}$). And $m$ of them are revealed uniformly at random in a balanced manner. Conditioned on the ground truth of clusters, indices of revealed nodes $\mathcal{R}$ and edges are independent. So, without loss of generality, we denote revealed labels $\Tilde{x}$ as follows:
\begin{equation}
    x_i=\begin{cases}
    1 \quad &i=1,2,\dots,\frac{n}{2},\\
    -1 \quad &i=\frac{n}{2},\frac{n}{2}+1,\dots,n;
    \end{cases}
    \quad
    \Tilde{x}_i=\begin{cases}
    1 \quad &i=1,2,\dots,\frac{m}{2},\\
    -1 \quad &i=\frac{n}{2},\frac{n}{2}+1,\dots,\frac{n+m}{2},\\
    0 \quad &\text{otherwise}.
    \end{cases} 
\end{equation}

We have shown that the entry value of the optimizer $X$ can be interpreted as an affinity metric among nodes. Moreover, we have $X_{ij} \in [-1, 1],~\forall~i,j$. It is natural to force the optimizer to have large entry values for those vertex pairs in which we have high confidence to be in the same community and vice versa. Therefore, we propose the CSDP approach to integrate the information provided by the semi-supervised approach. If node i and node j are revealed to have the same label, we add the constraint $X_{ij} = 1$ to the optimization model. If they are revealed to have the opposite labels, we add $X_{ij} = -1$. Formally, the CSDP is defined as 
\begin{equation}\CSDP(M_{n\times n}) = \max\{\la M,X\ra: X \in \text{elliptope}_n,~ X_{ij} = x_i \cdot x_j ~\forall i,j \in \mathcal{R}\}\end{equation}
where $\mathcal{R}$ denotes the collection of revealed nodes. After reordering the indices, we can assume it as $\{1,2,\dots,\frac{m}{2}\} \bigcup \{\frac{n}{2}, \frac{n}{2} + 1, \dots, \frac{n + m}{2}\}$. It is worth noting that the optimization remains a positive semidefinite programming problem, which can be solved efficiently, for example by interior point methods \citep{Alizadeh1995InteriorPM}.

Then let $\mathcal{S}^{n-1} \coloneqq \{v\in \mathbb{R}^n: \|v\|_2 = 1\}$ be the unit $(n-1)$-sphere and $\sigma = (\sigma_1,\sigma_2,\dots,\sigma_n) \in (\mathcal{S}^{n-1})^n$. Consider the CSDP in the form of Burer-Monteiro factorization. We have the following identities:
\begin{equation}
\SDP(M) = \max\left\{\sum_{i,j = 1}^n M_{ij}\la\sigma_i,\sigma_j \ra: \sigma_i \in \mathcal{S}^{n-1} ~\forall i \in [n]\right\},
\end{equation}
\begin{equation}
\CSDP(M) = \max_{\sigma \in (\mathcal{S}^{n-1})^n}\left\{\sum_{i,j = 1}^n M_{ij}\la\sigma_i,\sigma_j \ra: \sigma_i^\top\sigma_j = x_ix_j ~\forall i,j \in \mathcal{R}\right\}\end{equation}
\begin{equation}\label{eqn:CSDP decomp}
= \max_{\sigma \in (\mathcal{S}^{n-1})^n}
\left\{ \sum_{i,j \in [n]\setminus\mathcal{R}} M_{ij} \sigma_i^\top \sigma_j
+ \sum_{i,j \in \mathcal{R}} M_{ij} x_ix_j
+ 2\sum_{i\in \mathcal{R}}\sum_{j\in [n]\setminus\mathcal{R}} x_i M_{ij}\sigma_0^\top\sigma_j \right\}    
\end{equation}
where $\sigma_0 \equiv x_i\sigma_i$, $\forall i \in \mathcal{R}$. Now one can consider an alternative matrix with a special margin denoting the algebraic sum of the blocks from $M$ that are associated with $\mathcal{R}$. We define $M^\text{agg}$ to be the $(n-m+1) \times (n-m+1)$ symmetric matrix indexed from 0 that
\begin{align}
M^\text{agg}_{00} &= \sum_{i,j \in \mathcal{R}} M_{ij} x_ix_j,\\
M^\text{agg}_{0j} &= \sum_{i\in \mathcal{R}} x_i M_{i,j+\frac{m}{2}}\qquad\forall j \in [\frac{n}{2} - \frac{m}{2}],\\
M^\text{agg}_{0j} &= \sum_{i\in \mathcal{R}} x_i M_{i,j+m}\qquad\forall j \in [n - m]\setminus[\frac{n}{2} - \frac{m}{2}],\\
M^\text{agg}_{ij} &= M_{i+\frac{m}{2},j+\frac{m}{2}}\qquad\forall i,j \in [\frac{n}{2} - \frac{m}{2}],\\
M^\text{agg}_{ij} &= M_{i+m,j+m}\qquad\forall i,j \in [n - m]\setminus[\frac{n}{2} - \frac{m}{2}].
\end{align}
Essentially, we aggregate the rows and columns related to revealed vertices according to their communities into the $0$-th row and column and reindex the matrix. It introduces spikiness to the underlying matrix,
\begin{equation}M^\text{agg}=
\left(\begin{array}{@{}c|c@{}}
  \sum_{i,j \in \mathcal{R}} M_{ij} x_ix_j
  & \begin{matrix}
  M^\text{agg}_{01} & M^\text{agg}_{02} & \cdots & M^\text{agg}_{0,n-m}
  \end{matrix} \\
\hline
  \begin{matrix}
    M^\text{agg}_{01} \\ M^\text{agg}_{02} \\ \vdots \\ M^\text{agg}_{0,n-m}
  \end{matrix} &
  M_\mathcal{R^\complement}
\end{array}\right).
\end{equation}
Although \citep{montanari2015semidefinite} employed a rather different technique to study SDPs, they also noticed that the critical change comes with such built-in structures, where the authors state "We expect the phase transition in $\SDP(\lambda vv^\top + W)/n$ to depend---in general---on the vector $v$, and in particular on how ‘spiky’ this is".

Combining the transformed input matrix with equation~\eqref{eqn:CSDP decomp}, we conclude that CSDP is indeed an SDP regarding $M^\text{agg}$,
\begin{align} \label{eqn:CSDP}
\CSDP(M) &= \max_{\substack{\sigma_i \in \mathcal{S}^{n-m}\\ i = 0,1,\dots,n-m}}
\left\{ \sum_{i,j \in [n - m]} M^\text{agg}_{ij} \sigma_i^\top \sigma_j
+ M^\text{agg}_{00}
+ 2\sum_{j\in [n - m]} M^\text{agg}_{0j}\sigma_0^\top\sigma_j \right\}\\
&= \SDP(M^\text{agg}).
\end{align}

\begin{lemma} Let $M_{\mathcal{R}^\complement}$ be the principle submatrix of $M$ obtained by removing the rows and columns associated with $\mathcal{R}$. The following inequalities hold,
    \begin{equation}\SDP(M_{\mathcal{R}^\complement}) \leq \CSDP(M) - M^\text{agg}_{00}.\end{equation}
\end{lemma}
\begin{proof}
Let $X^*$ be the optimizer of $\SDP(M_{\mathcal{R}^\complement})$. Define its $(n - m + 1) \times (n - m + 1)$ extension $\hat{X^*}$ as 
\begin{equation}
\hat{X}^*_{ij} = 
\begin{cases}
1 & i=j=0,\\
0 &~ i \in [n-m],~ j =0,\\
0 &~ j \in [n-m],~ i = 0,\\
X^*_{ij} &~\text{otherwise}.\\
\end{cases}
\end{equation}
Due to the identity from above and the fact that $\hat{X^*}\in \text{elliptope}_{n-m+1}$ is feasible, we can conclude that
\begin{equation}\CSDP(M) = \SDP(M^\text{agg}) \geq \la \hat{X^*}, M^\text{agg} \ra = \SDP(M_{\mathcal{R}^\complement}) + M^\text{agg}_{00}.\end{equation}
\end{proof}

So far, all the results are deterministic, $M$ can be an arbitrary symmetric matrix, and $\mathcal{R}$ can be any balanced index set. Next, we will consider $M = \tilde{A} \coloneqq  A - \frac{d}{n} \mathbf{1}\mathbf{1}^\top$ to study CSDPs on probabilistic models.
\begin{remark}
As shown in the Lemma \ref{lemma:trueLabel}, $\tilde{A}^\text{agg}_{00} \geq m\cdot\frac{a-b}{2} \geq 0$ with high probability. By definition, we have $\CSDP(\tilde{A}) \leq \SDP(\tilde{A})$.
So, with high probability,
\begin{equation}\label{eqn: sandwich}
\SDP(\tilde{A}_{\mathcal{R}^\complement}) \leq \CSDP(\tilde{A}) \leq \SDP(\tilde{A}).
\end{equation}
The CSDP always lies in between the SDPs of the original adjacency matrix and the submatrix of unrevealed vertices. Moreover, if $\tilde{A} \sim \G(n,\frac{a}{n},\frac{b}{n})$, we have $ \tilde{A}_{\mathcal{R}^\complement}\sim \G(n - m,\frac{a(1-\rho)}{n - m},\frac{b(1-\rho)}{n - m})$. It is worth mentioning that although $\tilde{A}_{\mathcal{R}^\complement}$ is just a submatrix of the original centered adjacency matrix, its probabilistic distribution as a random matrix is not simply changed from $n$ nodes to $n-m$ nodes. The edge probability parameters are also changed by a factor of $(1-\rho)$. It leads to some technical challenges, which we are going to handle later. But intuitively, from the asymptotic behavior of SDP, we can derive a rough understanding of CSDP as $n\to \infty$. Recall that the phase transition theory tells us that when SNR $\leq$ 1, the optimal value of SDP for SBM will not be large enough to distinguish from the optimal value of SDP for ERM. Therefore, the order of the above quantities from inequality \eqref{eqn: sandwich} suggests that semi-supervised SDP can not help to increase the statistics associated with SBM. The best one can hope for is that it will make the statistics associated with ERM smaller by a factor depending on $\rho$. This turns out to be enough for community detection.
\end{remark}

Recall the community detection problem can be formalized as a binary hypothesis testing problem, whereby we want to determine, with a high probability of success, whether the random graph under consideration has a community structure or not. As discussed in Section \ref{summary}, we introduce semi-supervised learning to the problem by revealing a part of the labels involved in the random graph's generating process. Namely, if the labels associated with a graph $G$ over $n$ vertices are denoted as $x$, we choose $m$ of them uniformly at random and denote the index set by $\mathcal{R}$, such that $\sum_{i\in\mathcal{R}} x_i = 0$.

Given a realization of the random graph $G$ \textit{and the revealed labels $x_\mathcal{R}$}, we want to decide which of the following holds,
\begin{itemize}[leftmargin=1in]
    \item[Hypothesis 0:] $(G,x_\mathcal{R}) \sim \G(n,\frac{d}{n}, \rho)$ is an Erdős–Rényi random graph with edge probability $\frac{d}{n}$, $d = \frac{a+b}{2}$ and reveal ratio $\rho$. We denote the corresponding distribution over graphs by $\prob_0$.
    \item[Hypothesis 1:] $(G,x_\mathcal{R}) \sim \G(n, \frac{a}{n}, \frac{b}{n}, \rho)$ is a planted bisection random graph with edge probabilities $(\frac{a}{n}, \frac{b}{n})$ and reveal ratio $\rho$. We denote the corresponding distribution over graphs by $\prob_1$.
\end{itemize}
A statistical test $T$ is a function defined on the graphs and revealed labels with range $\{0,1\}$. It succeeds with a high probability if 
\begin{equation}\prob_0(T(G, x_\mathcal{R}) = 1) + \prob_1(T(G, x_\mathcal{R}) = 0) \to 0 \quad (n\to\infty).\end{equation}

Note that this is indeed a generalization of the unsupervised community detection. Simply looking at the labels, the two models are indistinguishable. What characterizes their difference is the probabilistic law of how edges are generated, i.e.,\ whether there is a cluster structure. The revealed labels serve as an enhancement of the graph observed. 
The phase transition theory says that under the unsupervised setting (the special case when $\rho = 0$), no test can succeed with high probability when SNR $\leq 1$, or equivalently, $a-b \leq \sqrt{2(a+b)}$. While if $\text{SNR}> 1$, several polynomially computable tests are developed. The SDP-based test is nearly optimal, in the sense that it requires
\begin{equation}\frac{a-b}{\sqrt{2(a+b)}}\geq 1+\epsilon(d)\end{equation}
where $\epsilon(d) \to 0$ as $d\to \infty$. It is believed to be the best that SDPs can reach. As the monotone-robustness study suggests \citep{Moitra2016HowRA}, this gap may be necessary, since SDP is indeed solving a harder problem where no algorithm can approach the threshold. However, we are going to see that when $\rho$ is sufficiently large, SDPs can not only reach but cross the threshold.

\subsection{Semi-Supervised Detectability}
With the problem and algorithm defined clearly, we are ready to prove that SBM and ERM can be consistently distinguished in the semi-supervised setting. We take a `divide and conquer' approach to establish an upper bound of CSDP on ERM, while we bound the CSDP on SBM from below with a witness that consists of the ground truth of labels, $X = xx^\tp$.

\begin{lemma} \label{lemma:trueLabel}
    Let $(A, x)$ obey the planted bisection model $G(n,\frac{a}{n}, \frac{b}{n})$ and denote $\la xx^\top, \tilde{A} \ra$ as $Y$. Then, for any $\epsilon > 0 $, we have $Y/n \in [\frac{a-b}{2} - \epsilon, \frac{a-b}{2} + \epsilon]$ with probability converging to one as $n\to \infty$.
\end{lemma}
\begin{proof}
\begin{align}
    Y&= \la xx^\top, \tilde{A} \ra = \la xx^\top, A \ra - \frac{d}{n}\la xx^\top, \1\1^\tp \ra \\
    &\overset{d}{=} 2\cdot\left[\text{Bin}\left(\left(\frac{n}{2}\right)^2 - \frac{n}{2},~ \frac{a}{n}\right) - \text{Bin}\left(\left(\frac{n}{2}\right)^2,~ \frac{b}{n}\right)\right].
\end{align}
We have $\E Y = \frac{n}{2}(a-b) - a$ and 
\begin{equation}\V Y = 4\left(a\left(\frac{n}{4} - \frac{1}{2}\right)\left(1-\frac{a}{n}\right) + b~\frac{n}{4}\left(1-\frac{b}{n}\right)\right)\leq n(a+b).\end{equation}
Then Chebyshev's inequality implies that for any $\delta \in (0,1)$,
\begin{align}
&\prob\left(|Y - \frac{n}{2}(a-b) + a| \geq \sqrt{n(a+b)} \cdot n^{(1-\delta)/2}\right) \leq \frac{1}{n^{1-\delta}}\\
&\implies\quad \prob\left(|\frac{Y}{n} - \frac{a-b}{2} + \frac{a}{n}| \geq \frac{\sqrt{a+b}}{n^{\delta/2}}\right) \leq \frac{1}{n^{1-\delta}}.\\
\end{align}
Hence, for sufficiently large $n$, we have 
\begin{equation}\prob\left(\frac{Y}{n} \geq \frac{a-b}{2} + \epsilon\right) + \prob\left(\frac{Y}{n} \leq \frac{a-b}{2} - \epsilon\right) \leq \frac{1}{n^{1-\delta}}.\end{equation}
Therefore,
\begin{equation}\prob\left(\frac{Y}{n}\in [\frac{a-b}{2} - \epsilon, \frac{a-b}{2} + \epsilon]\right) \geq 1 - \frac{1}{n^{1-\delta}}.\end{equation}
\end{proof}

Besides bounding the outcomes on the SBM from below, this lemma can also be applied to the `all revealed blocks' to estimate $\tilde{A}^\text{agg}_{00}$, which is used several times throughout our proofs.

\begin{lemma}\label{lemma:sbmLower}
Let $G \sim \G(n,\frac{a}{n},\frac{b}{n})$, $d = \frac{a+b}{2}$ and $\tilde{A} = A - \frac{d}{n}\1\1^\tp$ be its centered adjacency matrix. Then for any $\epsilon>0$ and $\gamma > 0$, with probability at least $1 - \frac{1}{n^{1-\gamma}}$, for all $n\geq n_0(a,b,\epsilon,\gamma)$, we have
    \begin{equation}\CSDP(\tilde{A}) \geq n~(\frac{a-b}{2} - \epsilon).\end{equation}
\end{lemma}
\begin{proof}
We prove the lower bound by considering a witness of the constrained optimization problem. Notice that $xx^\tp$ is feasible for both SDP and CSDP, where $x$ is the label vector associated with $G$. Therefore,
\begin{equation}
    \CSDP(\tilde{A}) \geq \la xx^\tp, \tilde{A} \ra.
\end{equation}

Then, we can apply Lemma \ref{lemma:trueLabel} to get the result.
\end{proof}
This result holds for any $\text{SNR} >0$ and suggests the following test for the semi-supervised community detection problem:
\begin{equation}
    T(G, x_\mathcal{R};\Delta) = \begin{cases}
    1 \quad\quad& \text{if~} \CSDP(\tilde{A}) \geq n[(a-b)/2 - \Delta],\\
    0 \quad\quad& \text{otherwise}.
    \end{cases}
\end{equation}

The following lemma bounds the CSDP of ERM from above. Intuitively, the contribution of blocks of the adjacency matrix, where columns or rows are associated with revealed nodes, concentrates well around zero. So the `effective dimension' of the SDP is reduced, hence the optimal value.
\begin{lemma}[Theorem 1, \citep{montanari2015semidefinite}. Reformulated.]\label{lemma:SDP}
Let $G \sim \G(n,\frac{d}{n})$ and $\tilde{A} = A - \frac{d}{n}\1\1^\tp$ be its centered adjacency matrix. There exists absolute constants $C$ and $d_0 > 1$ such that if $d \geq d_0 $, then with high probability,
    \begin{equation}\frac{1}{n\sqrt{d}}\SDP(\tilde{A}) \leq 2 + \frac{C \log d}{d^{1/10}}.\end{equation}
\end{lemma}
This result is rigorously derived with profound insights from mathematical physics. However, there is an implicit condition on the average degree $d$ in the proof. It is common to assume at least $d>1$ in the literature concerning unsupervised clustering because otherwise, the graph has no giant component, not to mention reconstruction, as discussed in Section \ref{topo}. But our approach leads to the subgraph with a possibly small effective average degree. Moreover, we do not want to be limited by the topology structure, although it is indeed a fundamental limit in the unsupervised setting. Theorem \ref{thm:CSDP} shows that semi-supervised SDPs are capable of integrating those sublinear components. To achieve that we resort to Grothendieck's inequality and carry out the analysis without assumption on $d$.

\begin{theorem}[Grothendieck's inequality \citep{Grothendieck1996RsumDL}]
Let $M$ be a $n\times n$ real matrix. If for any $s, t \in \{-1,1\}^n$,
\begin{equation}\label{eqn:infty21norm}
\big| \sum_{i,j} M_{ij}s_i t_j\big| \leq 1.
\end{equation}
Then for all vectors $X_i, Y_i \in \{x \in \mathbb{R}^n: \|x\|_2 \leq 1\},~ i = 1,2,\dots,n$, we have
\begin{equation}\label{Grothendieck}
\big|\sum_{i,j} M_{ij} \la X_i, Y_j\ra\big| \leq K_\text{G}.
\end{equation}
\end{theorem}

$K_\text{G}$ is an absolute constant called Grothendieck's constant. 
The inequality was initially proved as a fundamental tool in Functional Analysis. In this paper, we focus exclusively on the above matrix version in $\mathbb{R}^n$ and consider the following suboptimal estimate derived by \citep{Braverman2011TheGC},
\begin{equation}
K_\text{G} < \frac{\pi}{2\log(1+\sqrt{2})} \leq 1.78.
\end{equation}

Note that if we restrict the vectors $X_i$'s and $Y_i$'s to the unit sphere $\mathcal{S}^{n-1}$, the inequality is still true as any inequality valid for a set also holds for its subset. Since $s$ and $t$ are arbitrary, the left-hand side of inequality \eqref{eqn:infty21norm} is indeed the $\ell_\infty \to \ell_1$ norm of matrix $M$, which is
\begin{equation}
    \|M\|_{\infty \to 1} = \max_{\|x\|_\infty\leq 1}\|Mx\|_1 = \max_{s,t\in \{-1,1\}^n} s^\top M t = \max_{s,t\in \{-1,1\}^n}\big| \sum_{i,j} M_{ij}s_i t_j\big|.
\end{equation}
This norm is also known as the cut norm,
whose importance in algorithmic problems is well understood in the theoretical computer science community. With the elliptope definition of SDP from equation \eqref{eqn:SDP_} and the consequential factorization of $X$ in equation \eqref{BMF}, we can rewrite the theorem in the following matrix form.
\begin{lemma}\label{lemma:GI}
For arbitrary matrix $M \in \mathbb{R}^{n\times n}$, we have
\begin{equation}
    \SDP(M) \leq \max_{X \in \text{elliptope}_n} \big| \la M, X \ra\big| \leq K_\text{G} \|M\|_{\infty \to 1}.
\end{equation}
\end{lemma}

Next, we use Bernstein's inequality to establish a probabilistic bound on the cut norm of $A - \E A$ where $A$ is the adjacency matrix of $\G(n,\frac{d}{n})$.

\begin{theorem}[Bernstein's inequality \citep{Bernstein}]
Let $\{X_i\}_{i = 1}^n$ be independent random variables such that $\E X_i = 0$ and $|X_i| \leq M$ for any $i \in [n]$. Denote the average variance as $\sigma^2 = \frac{1}{n}\sum_{i=1}^n \V (X_i)$. Then for any $t \geq 0$,
\begin{equation}
    \prob \left(\frac{1}{n} \sum_{i=1}^n X_i > t \right) \leq \exp\left(-\frac{n t^2/2}{\sigma^2+\frac{Mt}{3}}\right).
\end{equation}
\end{theorem}

\begin{lemma}\label{lemma:cutNormBound}
Let $A$ be the adjacency matrix of an ERM, $\G(n,\frac{d}{n})$. Then, with probability at least $1-5^{-n+2}$,
\begin{equation}
\|A - \E A\|_{\infty \to 1} \leq 6(1+d)n.
\end{equation}
\end{lemma}

\begin{proof}
According to the identity from inequality \eqref{eqn:infty21norm}, we want to bound
\begin{align}\label{eqn:idSum}
    \|A - \E A\|_{\infty \to 1} &= \max_{s,t\in \{-1,1\}^n} \sum_{i,j}(A - \E A)_{ij}~s_i t_j\\
    &= \max_{s,t\in \{-1,1\}^n} \sum_{i<j} (A - \E A)_{ij}~(s_i t_j + s_j t_i).
\end{align}

For fixed $s,t\in \{-1,1\}^n$, denote
\begin{equation}
X_{ij} = (A - \E A)_{ij}~(s_i t_j + s_j t_i) \quad (1 \leq i < j \leq n).
\end{equation}
Then we have $\E X_{ij} = 0$, $|X_{ij}| \leq 2$ and $\V (X_{ij}) \leq 4 \frac{d}{n}$ for any $i<j$. There are totally $n(n-1)/2$ of $\{X_{ij}\}$'s. And they are independent by the definition of ERM. So Bernstein's inequality implies
\begin{equation}
\prob\left(\frac{2}{n(n-1)}\sum_{i<j}X_{ij}>t\right) \leq \exp\left(-\frac{n(n-1) t^2/4}{\frac{4d}{n}+\frac{2t}{3}}\right).
\end{equation}
Let $t = 12(1+d)/n$, which guarantees $4d/n+2t/3 < t$. Hence,
\begin{equation}
\prob\left(\sum_{i<j}X_{ij}>6(1+d)n\right) \leq \exp\left(-3(n-1)\right).
\end{equation}

Apply the union bound to all $2^{2n}$ possible $(s,t)$, we have
\begin{equation}
\prob \left(\max_{s,t\in \{-1,1\}^n} \sum_{i<j} (A - \E A)_{ij}~(s_i t_j + s_j t_i) >6(1+d)n \right) \leq 2^{2n} \cdot e^{-3(n-1)}.
\end{equation}
We conclude the proof with the identity of $\ell_\infty \to \ell_1$ norm and the fact that  the right-hand side of the above inequality is less than $5^{-n+2}$.
\end{proof}

Since the distribution of each entry in the matrix changes as $n\to \infty$, we now develop a slightly generalized version of the weak law of large numbers fitting for our purpose. We use the superscripts to explicitly denote dependence on $n$.

\begin{lemma} \label{lemma:WLLN}
For any $n$, let $\{X_i^{(n)}\}_{i = 1}^n$ be a collection of independent random variables. Assume there exist universal constants $\mu$ and $\sigma$, such that $\E X^{(n)}_i \leq \mu < \infty$ and $\V (X^{(n)}_i) \leq \sigma^2 < \infty$ for any $n \in \mathbb{N}$ and $i \leq n$. If we denote the sample mean as
\begin{equation}
\bar{X}^{(n)} = \frac{X_1^{(n)} + X_2^{(n)}+\dots+X_n^{(n)}}{n},
\end{equation}
then, for any $\epsilon > 0$,
\begin{equation}
\prob \big(\bar{X}^{(n)} \geq \mu + \epsilon\big) \to 0\quad \text{as}\quad n\to\infty.
\end{equation}
\end{lemma}

\begin{proof}
For any $n\in\mathbb{N}$, we have
\begin{align}
    \V (\bar{X}^{(n)}) &= \frac{1}{n^2} \V (X_1^{(n)} + X_2^{(n)}+\dots+X_n^{(n)}) \\
    &= \frac{\sum_{i = 1}^n \V(X_i^{(n)})}{n^2} & \text{(by independence)}\\
    &\leq \sigma^2/n. &\text{(by uniform boundedness)}
\end{align}
Then Chebyshev's inequality ensures
\begin{align}
   &\prob\big(|\bar{X}^{(n)} - \E \bar{X}^{(n)}|\geq \epsilon\big)\leq \frac{\sigma^2}{n \epsilon^2}\\
   \implies & \prob\big(\bar{X}^{(n)} \geq \frac{1}{n}\sum_{i = 1}^n \E X_i^{(n)} + \epsilon\big)\leq \frac{\sigma^2}{n \epsilon^2}\\
   \implies & \prob\big(\bar{X}^{(n)} \geq \mu + \epsilon\big)\leq \frac{\sigma^2}{n \epsilon^2}.
\end{align}
\end{proof}

\begin{remark}
Compared with a standard large deviation theory, this result allows the random variables to not be identically distributed. And more importantly, the distributions can depend on $n$. Furthermore, the random variables associated with different $n$ are not necessary to be independent.
\end{remark}
\begin{lemma}\label{lemma:boundMargin}
Let $G \sim \G(n,\frac{d}{n})$, $x$ be the labels, $\mathcal{R}$ be the revealed indices and $\tilde{A} = A - \frac{d}{n}\1\1^\tp$ be its centered adjacency matrix. Define
\begin{equation}
B_{ij} =\begin{cases}
\sum_{i,j \in \mathcal{R}} \tilde{A}_{ij} x_ix_j \quad &i=j=0,\\
\sum_{k\in \mathcal{R}} x_k\tilde{A}_{kj} \quad & i = 0,~j \in [n]\setminus \mathcal{R},\\
\sum_{k\in \mathcal{R}} x_k\tilde{A}_{ik} \quad & j = 0,~i \in [n]\setminus \mathcal{R},\\
0 \quad & \text{otherwise}.
\end{cases}\end{equation}
Then for any $\epsilon >0$, with high probability,
\begin{equation}
\SDP(B)\leq 2dm(1-\frac{m}{n}) + (2n - m)\epsilon.
\end{equation}
\end{lemma}
\begin{proof}
Notice that for any feasible $X$ of the above optimization problem, we have $X\succeq0, X_{ii}=1 ~\forall i \in [n+1]$. So, for any $i,~j \in [n+1]$,
\begin{equation}
(\mathbf{e}_i \pm \mathbf{e}_j)^\tp X (\mathbf{e}_i \pm \mathbf{e}_j) = 2 \pm 2X_{ij} \geq 0 \quad \implies \quad |X_{ij}| \leq 1.
\end{equation}
Therefore,
\begin{align}
\SDP(B) &= \max\{\la B,X\ra: X \in \text{elliptope}_{n+1}\}\\
    &=B_{00} + 2\max\left\{\sum_{j\in [n]\setminus \mathcal{R}} B_{0j}X_{0j}: X \in \text{elliptope}_{n+1}\right\}\\
    &\leq B_{00} + 2\sum_{j\in [n]\setminus \mathcal{R}} |B_{0j}|.
\end{align}
Note that $\{B_{0j}: j\in [n]\setminus \mathcal{R}\}$'s are independent random variables. Moreover, if we let $B_1,~B_2$ be two independent binomial random variables with the same parameter $(\frac{m}{2}, \frac{d}{n})$ and denote their difference as $Z\coloneqq B_1-B_2$, we have
 $B_{0j} \overset{\text{d}}{=}Z$ for any $j\in [n]\setminus \mathcal{R}$ with $\E Z = 0$ and $\V Z \leq d\frac{m}{n}$.
 
Since $Z^2 \geq |Z|$, we have
\begin{align}
&\E |Z| \leq \E (Z^2) = \V Z \leq d\frac{m}{n},\\
&\V |Z| = E (Z^2) - (\E |Z|)^2\leq \V Z \leq d\frac{m}{n}.
\end{align}

Then Lemma \ref{lemma:WLLN} can be applied to 
\begin{equation}
\bar{X}^{(n)} \coloneqq \frac{\sum_{j\in [n]\setminus \mathcal{R}} |B_{0j}|}{n-m}.
\end{equation}
So, for any $\epsilon>0$, we have
\begin{equation}
\lim_{n\to \infty} \prob \left(\frac{1}{ n-m}\sum_{j\in [n]\setminus \mathcal{R}} |B_{0j}|>d\frac{m}{n} + \epsilon\right) = 0.
\end{equation}
Hence, $\sum_{j\in [n]\setminus \mathcal{R}} |B_{0j}|\leq (n-m)(d\frac{m}{n}+\epsilon)$ with high probability.

Lemma \ref{lemma:trueLabel} implies, with high probability,
\begin{equation}B_{00} \leq \epsilon m.\end{equation}

Combining the above results with the union bound completes the proof.
\end{proof}

Returning to the semi-supervised SDP, based on the notions from Section \ref{CSDP def}, we consider the following decomposition of the transformed input matrix $M^\text{agg}$ with the unrevealed part and revealed part as
\begin{equation}\label{eqn:decomp}
    M^\text{agg} = M^{(\mathcal{R}^\complement)} + M^{(\mathcal{R})}
\end{equation}
where we define
\begin{equation}
M^{(\mathcal{R})}_{ij} = 
\begin{cases}
M^\text{agg}_{ij} &\quad i = 0 \text{~or~} j = 0,\\
0 &\quad \text{otherwise}.
\end{cases}
\end{equation}

To prove the main result of semi-supervised SDP, we control the $M^{(\mathcal{R}^\complement)}$ part by Grothendieck's inequity and bound the contribution of $M^{(\mathcal{R})}$ with the generalized law of large numbers shown above.

\bigbreak

\begin{proof}[Proof of Theorem \ref{thm:CSDP}]
Notice that Lemma \ref{lemma:sbmLower} guarantees the test to succeed under the SBM. We only need to show, under ERM, 
\begin{equation}\label{ubgoal}
    \CSDP(\tilde{A}) < n[(a-b)/2 - \Delta] \quad \text{(w.h.p.)}.
\end{equation}
According to the identity from equation \eqref{eqn:CSDP}, we have
\begin{align}
    \CSDP(\tilde{A}) &= \SDP(\tilde{A}^\text{agg})\\
    &= \max\{\la \tilde{A}^\text{agg},X\ra: X \in \text{elliptope}_n\}\\
    &=\max\{\la \tilde{A}^{(\mathcal{R}^\complement)} + \tilde{A}^{(\mathcal{R})},X\ra: X \in \text{elliptope}_n\}\\
    &\leq \SDP(\tilde{A}^{(\mathcal{R}^\complement)}) + \SDP(\tilde{A}^{(\mathcal{R})}).
\end{align}

Recall that $\tilde{A}_{\mathcal{R}^\complement}$ is the principal submatrix of $\tilde{A}$ obtained by removing the rows and columns associated with $\mathcal{R}$. By definition, we have $\SDP(\tilde{A}_{\mathcal{R}^\complement}) = \SDP(\tilde{A}^{(\mathcal{R}^\complement)})$. Under the null hypothesis, $\tilde{A}_{\mathcal{R}^\complement}$ has the same distribution as the centered adjacency matrix associated with $\G(n-m,\frac{(1-\rho)d}{n-m})$. Also,
\begin{align}
    \SDP
    \left(\tilde{A}^{(\mathcal{R}^\complement)} \right) &= \SDP\left(A_{\mathcal{R}^\complement} - \E A_{\mathcal{R}^\complement} - \frac{(1-\rho)d}{n-m} I_{n-m}\right)\\
    &=\SDP\left(A_{\mathcal{R}^\complement} - \E A_{\mathcal{R}^\complement} \right) - (1-\rho)d.
\end{align}

Now we apply Grothendieck's inequality and Lemma \ref{lemma:cutNormBound}. With probability at least $1-5^{-(1-\rho)n+2}$,
\begin{equation}
    \SDP
    \left(\tilde{A}^{(\mathcal{R}^\complement)} \right) \leq 6K_\text{G}[1+(1-\rho)d](n-m) < 12(1+d)(1-\rho)n.
\end{equation}
Combining above estimations and the result from Lemma \ref{lemma:boundMargin} with $\epsilon = d(1-\rho)^2$, we have
\begin{equation}
    \frac{1}{n}\CSDP(\tilde{A}) \leq 14(1-\rho)(1+d) \quad \text{(w.h.p.)}.
\end{equation}

Hence, Equation \eqref{ubgoal} holds with $\Delta = (a-b)/40$ and $\rho \geq \rho_0 = 1 - \frac{a-b}{30(1+d)}$. We conclude, if $\frac{m}{n} \geq \rho _0$, 
\begin{equation}\prob_0 (T(G, x_\mathcal{R}) = 1) \to 0 \quad (n\to \infty).\end{equation}
\end{proof}

\begin{remark}
    If $\rho = 0$, CSDP is naturally reduced to SDP. Hence, it shares the same capability to solve the (unsupervised) community detection problem when $\text{SNR} > 1$ as stated in Theorem \ref{MS}. Although the analysis above cannot be directly generalized to a vanishing $\rho \to 0$ situation, CSDP provides a new perspective for further study on the optimality of SDP.  
\end{remark}

\begin{remark}
    We believe that it is possible to reduce the requirement on $\rho$ through more involved analysis. Empirically, it does not need to be close to 1 for a clear improvement in the result. For instance, our simulation in the next section shows that phase transition disappears with $20\%$ of the nodes revealed.
\end{remark}

\section{Numerical Experiments}\label{numexp}
We include some simulation results below. $\rho \in [0,1]$ is the ratio of revealed labels. Results associated with unsupervised SDPs are identified as $\rho = 0$. As discussed in Section \ref{Census method}, to make the comparison fair and keep the problem meaningful, all overlaps are restricted to the unrevealed labels.

\begin{figure}[htbp] 
    \centering\includegraphics[width=0.95\textwidth]{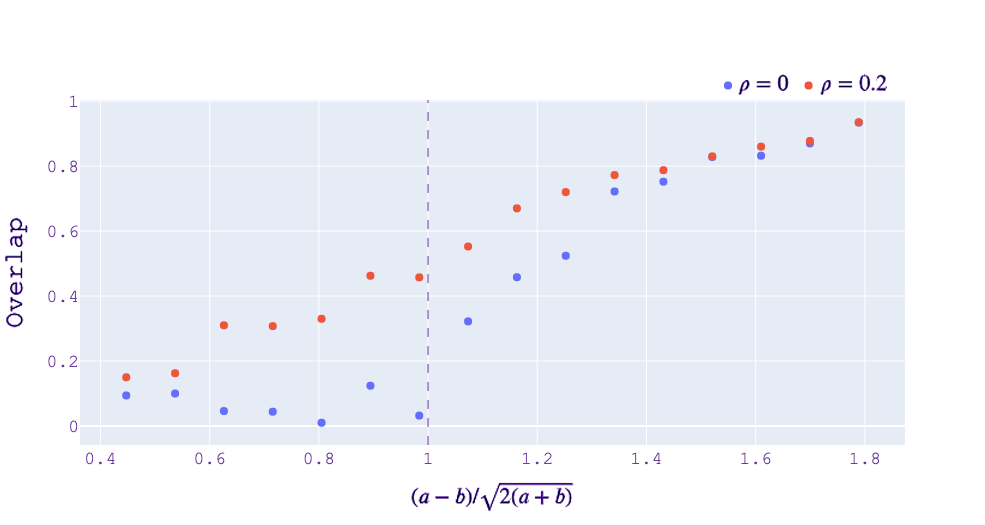}
    \caption{Disappearance of the phase transition.}
    \label{fig:disappear}
\end{figure}

Each point in Figure \ref{fig:disappear} represents one realization of a SBM with $n=1000$. The dashed line stands for the KS and information-theoretic threshold. The graphs are shared by both the unsupervised and the semi-supervised SDPs. Labels are identified by applying the k-means method to corresponding optimizers. 
% Notice that in the sparse regime, directly applying k-means to the adjacency matrix or its semi-supervised modification, i.e.,\ updating the entries associated with the relationships among revealed nodes by 1 or -1, will obviously get an overlap close to zero.
Overlaps of unsupervised SDP essentially drop down to zero on the left-hand side. In contrast, with $20\%$ of the labels revealed, the outcome of our constraint SDP algorithm goes down gradually as the SNR decreases and remains substantially greater than zero even when $\text{SNR}\leq 1$.

\begin{figure}[ht]
\begin{subfigure}{0.43\textwidth}
\centering\includegraphics[width=\textwidth]{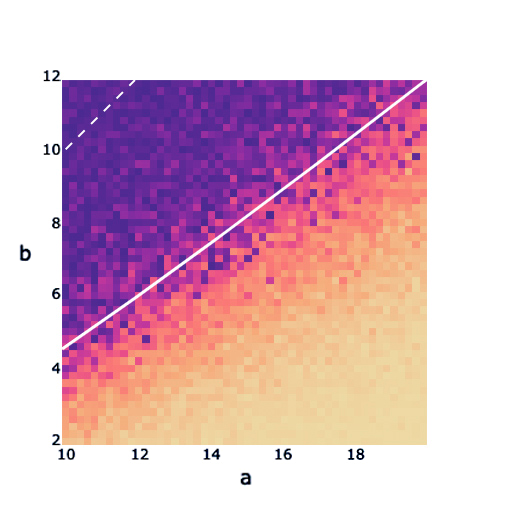}
\end{subfigure}
\hspace*{\fill}
\begin{subfigure}{0.43\textwidth}
\centering\includegraphics[width=\textwidth]{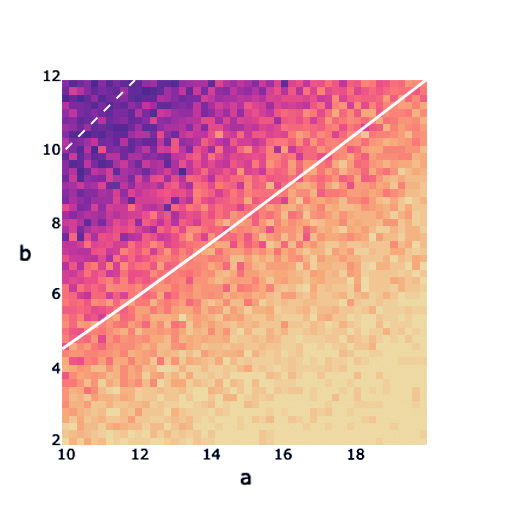}
\end{subfigure}
\hspace*{\fill}
\begin{subfigure}{0.1\textwidth}
\centering\includegraphics[width=\textwidth]{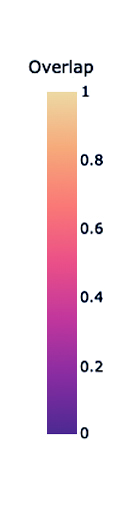}
\end{subfigure}
\hspace*{\fill} 
\caption{Overlap heatmaps of the unsupervised (left) and the semi-supervised (right) SDPs. The coordinates correspond to the model parameters $a$ and $b$. The solid line represents the KS and information-theoretic threshold. The dashed line corresponds to $a=b$.}
\label{fig:phase}
\end{figure}

\vspace{1cm}
Theorem \ref{KS} guarantees that the upper left corner of the left image will be completely dark as $n\to \infty$. But we see semi-supervised SDPs successfully `light up' the area between the two reference lines, see Figure~\ref{fig:phase}. Moreover, when $n$ is sufficiently large, there will be no pixel with a value of 0.

\begin{figure}[p]
\begin{subfigure}{0.48\textwidth}
\centering\includegraphics[width=\textwidth]{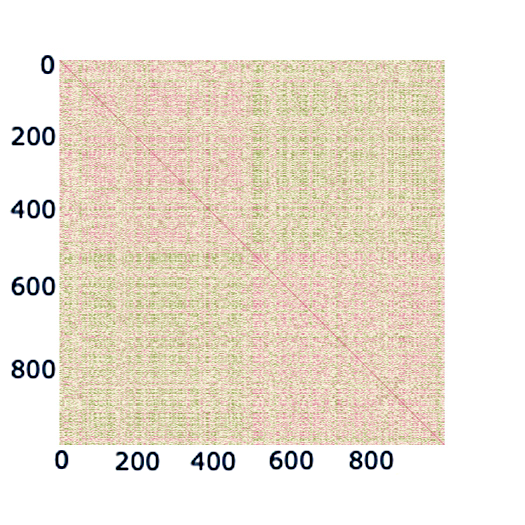}
\end{subfigure}
\hspace*{\fill}
\begin{subfigure}{0.48\textwidth}
\centering\includegraphics[width=\textwidth]{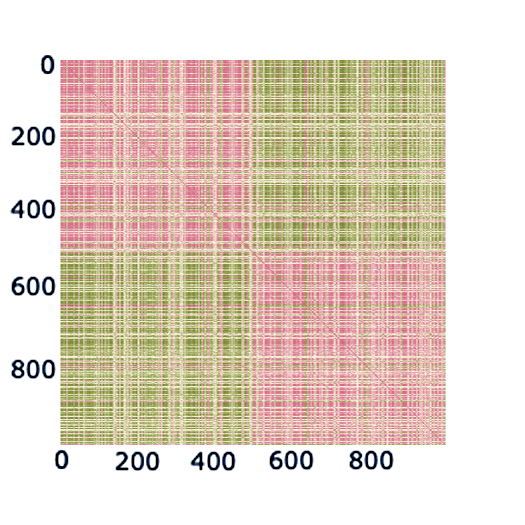}
\end{subfigure}
\hspace*{\fill}
\begin{subfigure}{0.48\textwidth}
\centering\includegraphics[width=\textwidth]{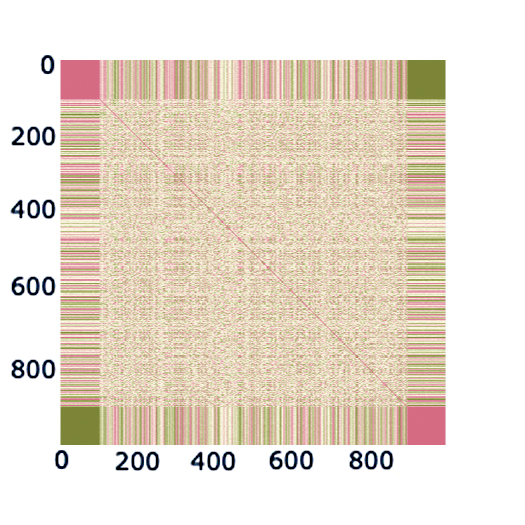}
\end{subfigure}
\hspace*{\fill}
\begin{subfigure}{0.48\textwidth}
\centering\includegraphics[width=0.28\textwidth]{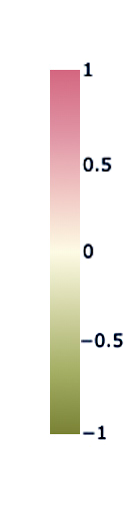}
\end{subfigure}
\hspace*{\fill} 
\caption{Visualization of the optimizer $X^*$. The upper row is concerned with one realization of the SBM $\G(1000, 12/1000, 5/1000)$, where the left image shows the value of optimizer for the unsupervised SDP and the right image is associated with the semi-supervised SDP with $\rho = 0.2$. The lower left image is the optimizer for one realization of the ERM of the same size with the associated average degree $d = 8.5$, indices of which are reordered such that the entries related to revealed labels are gathered in four corners. It could be understood as the situation of the null hypothesis we defined in Section \ref{CSDP def}.}
\label{fig:optX}
\end{figure}

Figure \ref{fig:optX} shows color-coded entry values of optimizer $X^*$ in different settings and suggests that the representing of underlying community structure is significantly enhanced by the semi-supervised approach, while no such structure is introduced if there should not be one.

\begin{figure}[p] 
    \text{SDP optimal value, when SNR is above KS/IT}\par\medskip
    \centering\includegraphics[width=0.8\textwidth]{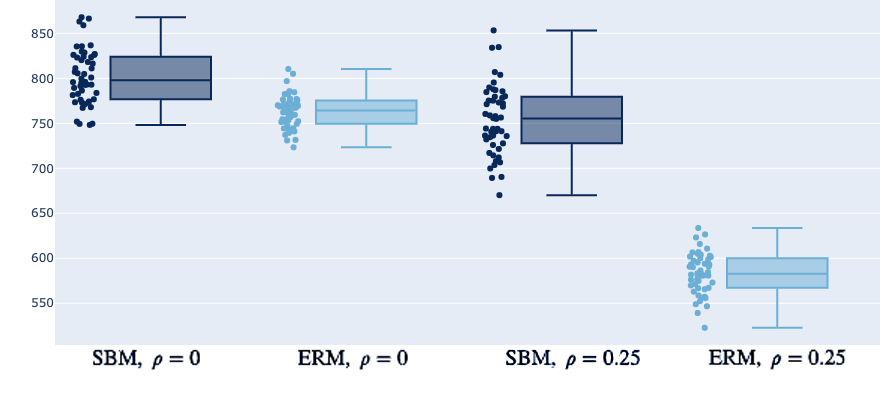}
    \caption{$a = 9,~b = 2 \quad (d = 5.5,~ \text{SNR} \approx 2.23)$}
    \label{fig:a9b2}
\end{figure}

\begin{figure}[p] 
    \text{SDP optimal value, when SNR is below KS/IT}\par\medskip
    \centering\includegraphics[width=0.8\textwidth]{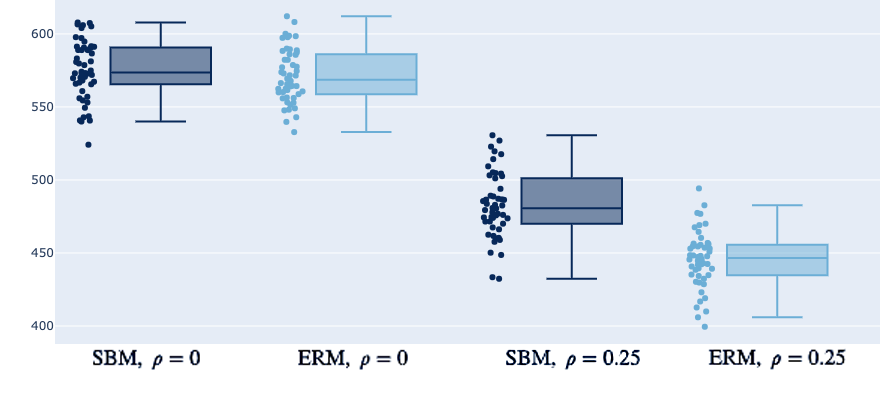}
    \caption{$a = 5,~b = 2 \quad (d = 3.5,~ \text{SNR} \approx 0.64)$}
    \label{fig:a5b2}
\end{figure}

To see how such a better representation leads to a successful test that is originally impossible, we consider the following simulations. We generate $50$ independent realizations of underlying random graphs ($n=200$) and compute their SDP values with and without the semi-supervised constraints ($\rho = 0.25$). Particularly, the parameters in Figure \ref{fig:a9b2} are chosen to have $\text{SNR}> 1$. The left two boxes imply that we can tell the difference between SBM and the ERM with the same average degree $d = (a+b)/2$. However, as in Figure \ref{fig:a5b2}, the vanilla SDP gives essentially the same result since the two models become contiguous if $\text{SNR}\leq 1$. As we have proved in Theorem \ref{thm:CSDP}, our semi-supervised SDP algorithm still manages to distinguish them by bringing down the optimal value of ERM more significantly when compared to its effect on SBM, which is confirmed by the right two boxes.

\newpage
\section{Conclusion}\label{conclusion}
The census method comes from the combinatorial perspective, while the CSDP is inspired by convex optimization research. Both algorithms are computationally efficient. The former has no requirement on the reveal ratio. The latter one is more practical and backward compatible with the unsupervised setting. By carefully integrating the revealed information with the observed graph structure, we can not only improve the performance of clustering algorithms but also resolve initially unsolvable problems. The fundamental changes brought by the semi-supervised approach let us cross the KS threshold, information-theoretical threshold, and even the topological limitation.

Our work provides a different angle to study stochastic models of networks and semidefinite programs. In real-world situations, it is almost always the case that we will have a certain fraction of samples being understood fairly well. So, an abstract model should be able to capture the existence of such knowledge instead of being blindly restricted to the unsupervised setting. Combining the universality of `revealed' information and the insight derived from our census method, it is arguable that the phase transitions, although mathematically beautiful, will never be an issue in practice. Our results on CSDPs, in turn, could be used to study SDPs, e.g.,\ prove or disprove that it can reach the phase transition threshold or the monotone-robustness threshold by a limiting process of $\rho \to 0$.

\if 0
Besides the mathematical curiosity, a major reason we study these foundational problems, e.g.,\ clustering on random graphs, is to develop better tools for realistic applications via theoretical guidance. We see such a promising direction of using the optimizer $X^*$ from CSDP as the propagation model $\hat{A}$ of GCN, see equation \ref{GCN}. One can view $X^*$ as a better representation of the node similarity, see for example Figure \ref{fig:optX}. The key idea behind GCN is exactly making similar nodes share the activation. Moreover, in a deep learning setting, the revealed labels are usually sufficient. CSDP will provide a learning objective justified graph representation, which not only contains more information about the underlying community structure but also auto-calibrates to the specific learning task. 
\fi

\acks{The authors acknowledge support from the National Science Foundation via grants NSF DMS-2027248, NSF CCF-1934568, NIH grants P41EB032840, R01HL16351, and DE-SC0023490.}

\newpage
\bibliography{SSC}
\end{document}